\documentclass[runningheads]{llncs}
\pdfoutput=1

\usepackage{graphicx}
\usepackage{amsmath,amssymb} 
\usepackage{color}
\usepackage{graphics}
\usepackage{hhline}
\usepackage{sidecap}
\usepackage{multirow}
\usepackage{array}
\usepackage{adjustbox}
\usepackage[ruled,vlined]{algorithm2e}
\usepackage[width=122mm,left=12mm,paperwidth=146mm,height=193mm,top=12mm,paperheight=217mm]{geometry}
\begin{document}

\pagestyle{headings}
\mainmatter

\def\etal{\emph{et al.}}
\def\ie{\emph{i.e.~}}
\def\eg{\emph{e.g.~}}
\newcommand{\no}[1]{}
\newcommand{\lbird}{L-Bird}
\newcommand{\lair}{L-Aircraft}
\newcommand{\llep}{L-Butterfly}
\newcommand{\ldog}{L-Dog}
\newcommand{\fgvc}{FGVC}
\newcommand{\fgvcgt}{FGVC-GT}
\newcommand{\cubgt}{CUB-GT}
\newcommand{\birdsnapgt}{Birdsnap-GT}
\newcommand{\doggt}{Dogs-GT}
\newcommand{\argmin}{\mathop{\arg\min}}
\newcommand{\argmax}{\mathop{\arg\max}}

\newcommand{\repeatthanks}{\textsuperscript{\thefootnote}}

\title{The Unreasonable Effectiveness of Noisy Data for Fine-Grained Recognition}

\titlerunning{The Unreasonable Effectiveness of Noisy Data for Fine-Grained Recognition}

\authorrunning{Krause \etal}

\author{
  Jonathan Krause\textsuperscript{1}\thanks{Work done while J. Krause was interning at Google} \quad Benjamin Sapp\textsuperscript{2}\thanks{Work done while B. Sapp and J. Philbin were at Google} \quad Andrew Howard\textsuperscript{3} \\
  Howard Zhou\textsuperscript{3} \quad Alexander Toshev\textsuperscript{3} \quad Tom Duerig\textsuperscript{3} \\
  James Philbin\textsuperscript{2}\repeatthanks  \quad Li Fei-Fei\textsuperscript{1} \\
  \textsuperscript{1}Stanford University \qquad \textsuperscript{2}Zoox \qquad \textsuperscript{3}Google\\
  {\tt\small \{jkrause,feifeili\}@cs.stanford.edu \quad \{bensapp,james\}@zoox.com} \\
  {\tt\small \{howarda,howardzhou,toshev,tduerig\}@google.com }
}

\institute{} 

\maketitle

\begin{abstract}
Current approaches for fine-grained recognition do the following: First, recruit experts to annotate a dataset of images, optionally also collecting more structured data in the form of part annotations and bounding boxes.
Second, train a model utilizing this data.
Toward the goal of solving fine-grained recognition, we introduce an alternative approach, leveraging free, noisy data from the web and simple, generic methods of recognition.
This approach has benefits in both performance and scalability.
We demonstrate its efficacy on four fine-grained datasets, greatly exceeding existing state of the art without the manual collection of even a single label, and furthermore show first results at scaling to more than 10,000 fine-grained categories.
Quantitatively, we achieve top-1 accuracies of $92.3\%$ on CUB-200-2011, $85.4\%$ on Birdsnap, $93.4\%$ on FGVC-Aircraft, and $80.8\%$ on Stanford Dogs without using their annotated training sets.
We compare our approach to an active learning approach for expanding fine-grained datasets.
\end{abstract}

\section{Introduction}
Fine-grained recognition refers to the task of distinguishing very similar categories, such as breeds of dogs~\cite{khosla2011novel,liu2012dog}, species of birds~\cite{wahcub2002011,horn2015,branson2014bird,bergbirdsnapcvpr2014}, or models of cars~\cite{yang2015large,krause20133d}.
Since its inception, great progress has been made, with accuracies on the popular CUB-200-2011 bird dataset~\cite{wahcub2002011} steadily increasing from 10.3\%~\cite{wahcub2002011} to 84.6\%~\cite{xu2015augmenting}.

\begin{figure}[t]
\centering
\includegraphics[width=0.45\linewidth]{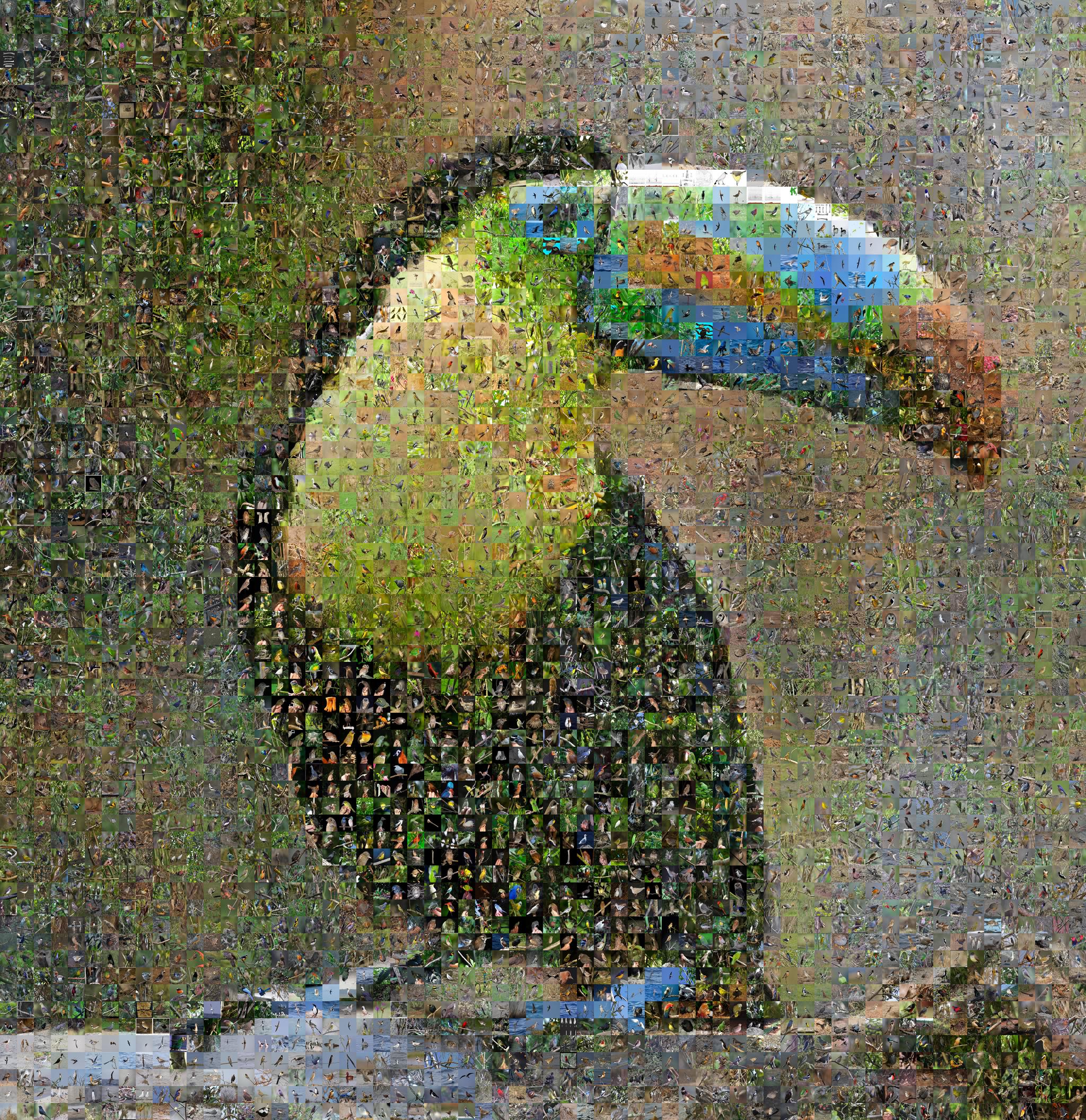}
\caption{
There are more than 14,000 species of birds in the world.
In this work we show that using noisy data from publicly-available online sources can not only improve recognition of categories in today's datasets, but also scale to very large numbers of fine-grained categories, which is extremely expensive with the traditional approach of manually collecting labels for fine-grained datasets.
Here we show 4,225 of the 10,982 categories recognized in this work.
}
\label{fig:pull}
\end{figure}

The predominant approach in fine-grained recognition today consists of two steps.
First, a dataset is collected.
Since fine-grained recognition is a task inherently difficult for humans, this typically requires either recruiting a team of experts~\cite{horn2015,maji13finegrained} or extensive crowd-sourcing pipelines~\cite{krause20133d,bergbirdsnapcvpr2014}. 
Second, a method for recognition is trained using these expert-annotated labels, possibly also requiring additional annotations in the form of parts, attributes, or relationships~\cite{zhang2014part,jaderberg2015spatial,lin2015bilinear,branson2014bird}.
While methods following this approach have shown some success~\cite{branson2014bird,zhang2014part,lin2015bilinear,krause2014icpr}, their performance and scalability is constrained by the paucity of data available due to these limitations.
With this traditional approach it is prohibitive to scale up to all 14,000 species of birds in the world (Fig.~\ref{fig:pull}), 278,000 species of butterflies and moths, or 941,000 species of insects~\cite{Hinchliff18092015}.

In this paper, we show that it is possible to train effective models of fine-grained recognition using noisy data from the web and simple, generic methods of recognition~\cite{szegedy2015rethinking,szegedy2014going}.
We demonstrate recognition abilities greatly exceeding current state of the art methods, achieving top-1 accuracies of $92.3\%$ on CUB-200-2011~\cite{wahcub2002011}, $85.4\%$ on Birdsnap~\cite{bergbirdsnapcvpr2014}, $93.4\%$ on FGVC-Aircraft~\cite{maji13finegrained}, and $80.8\%$ on Stanford Dogs~\cite{khosla2011novel} \emph{without using a single manually-annotated training label from the respective datasets}.
On CUB, this is nearly at the level of human experts~\cite{branson2014ignorant,horn2015}.
Building upon this, we scale up the number of fine-grained classes recognized, reporting first results on over 10,000 species of birds and 14,000 species of butterflies and moths.

The rest of this paper proceeds as follows:
After an overview of related work in Sec.~\ref{sec:related}, we provide an analysis of publicly-available noisy data for fine-grained recognition in Sec.~\ref{sec:data}, analyzing its quantity and quality.
We describe a more traditional active learning approach for obtaining larger quantities of fine-grained data in Sec.~\ref{sec:active-learning}, which serves as a comparison to purely using noisy data.
We present extensive experiments in Sec.~\ref{sec:experiments}, and conclude with discussion in Sec.~\ref{sec:conclusion}.

\section{Related Work}
\label{sec:related}

\subsubsection{Fine-Grained Recognition.}
The majority of research in fine-grained recognition has focused on developing improved models for classification~\cite{angelova2013image,berg2013poof,branson2014bird,chai2011bicos,chai2012tricos,chai2013symbiotic,deng2013fine,duan2012discovering,farrell2011birdlets,gavves2013fine,gavves2014local,goering2014nonparametric,krause2014icpr,krause2015fine,lin2015bilinear,liu2012dog,nilsback2006visual,pu2014looks,shih2015part,simon14pdd,simon2015neural,xiao2015application,xie2014hyper,xu2015augmenting,yang2012unsupervised,yao2011combining,yao2012codebook,zhang2012pose,zhang2013deformable,zhang2014part,zhang2015weakly}.
While these works have made great progress in modeling fine-grained categories given the limited data available, very few works have considered the impact of that data~\cite{xu2015augmenting,xie2014hyper,horn2015}.
Xu \etal~\cite{xu2015augmenting} augment datasets annotated with category labels and parts with web images in a multiple instance learning framework, and Xie \etal~\cite{xie2014hyper} do multitask training, where one task uses a ground truth fine-grained dataset and the other does not require fine-grained labels.
While both of these methods have shown that augmenting fine-grained datasets with additional data can help, in our work we present results which completely forgo the use of any curated ground truth dataset.
In one experiment hinting at the use of noisy data, Van Horn \etal~\cite{horn2015} show the possibility of learning 40 bird classes from Flickr images.
Our work validates and extends this idea, using similar intuition to significantly improve performance on existing fine-grained datasets and scale fine-grained recognition to over ten thousand categories, which we believe is necessary in order to fully explore the research direction.

Considerable work has also gone into the challenging task of curating fine-grained datasets~\cite{bergbirdsnapcvpr2014,horn2015,khosla2011novel,krause20133d,kumar2012leafsnap,mahendran14understanding,welinderetal2010,wahcub2002011,yang2015large} and developing interactive methods for recognition with a human in the loop~\cite{branson2014ignorant,wah2011multiclass,wah2013attribute,wah2014similarity}.
While these works have demonstrated effective strategies for collecting images of fine-grained categories, their scalability is ultimately limited by the requirement of manual annotation.
Our work provides an alternative to these approaches.

\subsubsection{Learning from Noisy Data.}
Our work is also inspired by methods that propose to learn from web data~\cite{divvala2014learning,chen2015webly,chen2013neil,schroff2011harvesting,li2010optimol,fergus2010learning} or reason about label noise~\cite{mnih2012learning,xiao2015learning,horn2015,sukhbaatar2014learning,reed2014training}.
Works that use web data typically focus on detection and classification of a set of coarse-grained categories, but have not yet examined the fine-grained setting.
Methods that reason about label noise have been divided in their results: some have shown that reasoning about label noise can have a substantial effect on recognition performance~\cite{xiao2015application}, while others demonstrate little change from reducing the noise level or having a noise-aware model~\cite{sukhbaatar2014learning,reed2014training,horn2015}.
In our work, we demonstrate that noisy data can be surprisingly effective for fine-grained recognition, providing evidence in support of the latter hypothesis.

\section{Noisy Fine-Grained Data}
\label{sec:data}
In this section we provide an analysis of the imagery publicly available for fine-grained recognition, which we collect via web search.\footnote{Google image search: \url{http://images.google.com}}
We describe its quantity, distribution, and levels of noise, reporting each on multiple fine-grained domains.

\subsection{Categories}
\label{sec:data_categories}
We consider four domains of fine-grained categories: birds, aircraft, Lepidoptera (a taxonomic order including butterflies and moths), and dogs.
For birds and Lepidoptera, we obtained lists of fine-grained categories from Wikipedia, resulting in 10,982 species of birds and 14,553 species of Lepidoptera, denoted \lbird{} (``Large Bird'') and \llep{}.
For aircraft, we assembled a list of 409 types of aircraft by hand (including aircraft in the FGVC-Aircraft~\cite{maji13finegrained} dataset, abbreviated \fgvc{}).
For dogs, we combine the 120 dog breeds in Stanford Dogs~\cite{khosla2011novel} with 395 other categories to obtain the 515-category L-Dog.
We evaluate on two other fine-grained datasets in addition to \fgvc{} and Stanford Dogs: CUB-200-2011~\cite{wahcub2002011} and Birdsnap~\cite{bergbirdsnapcvpr2014}, for a total of four evaluation datasets.
CUB and Birdsnap include 200 and 500 species of common birds, respectively, \fgvc{} has 100 aircraft variants, and Stanford Dogs contains 120 breeds of dogs.
In this section we focus our analysis on the categories in \lbird{}, \llep{}, and \lair{} in addition to the categories in their evaluation datasets.

\subsection{Images from the Web}
\label{sec:data_imagery}
We obtain imagery via Google image search results, using all returned images as images for a given category.
For \lbird{} and \llep{}, queries are for the scientific name of the category, and for \lair{} and L-Dog queries are simply for the category name (\eg ``Boeing 737-200'' or ``Pembroke Welsh Corgi'').


\begin{figure}[t]
\centering
\includegraphics[width=0.23\linewidth]{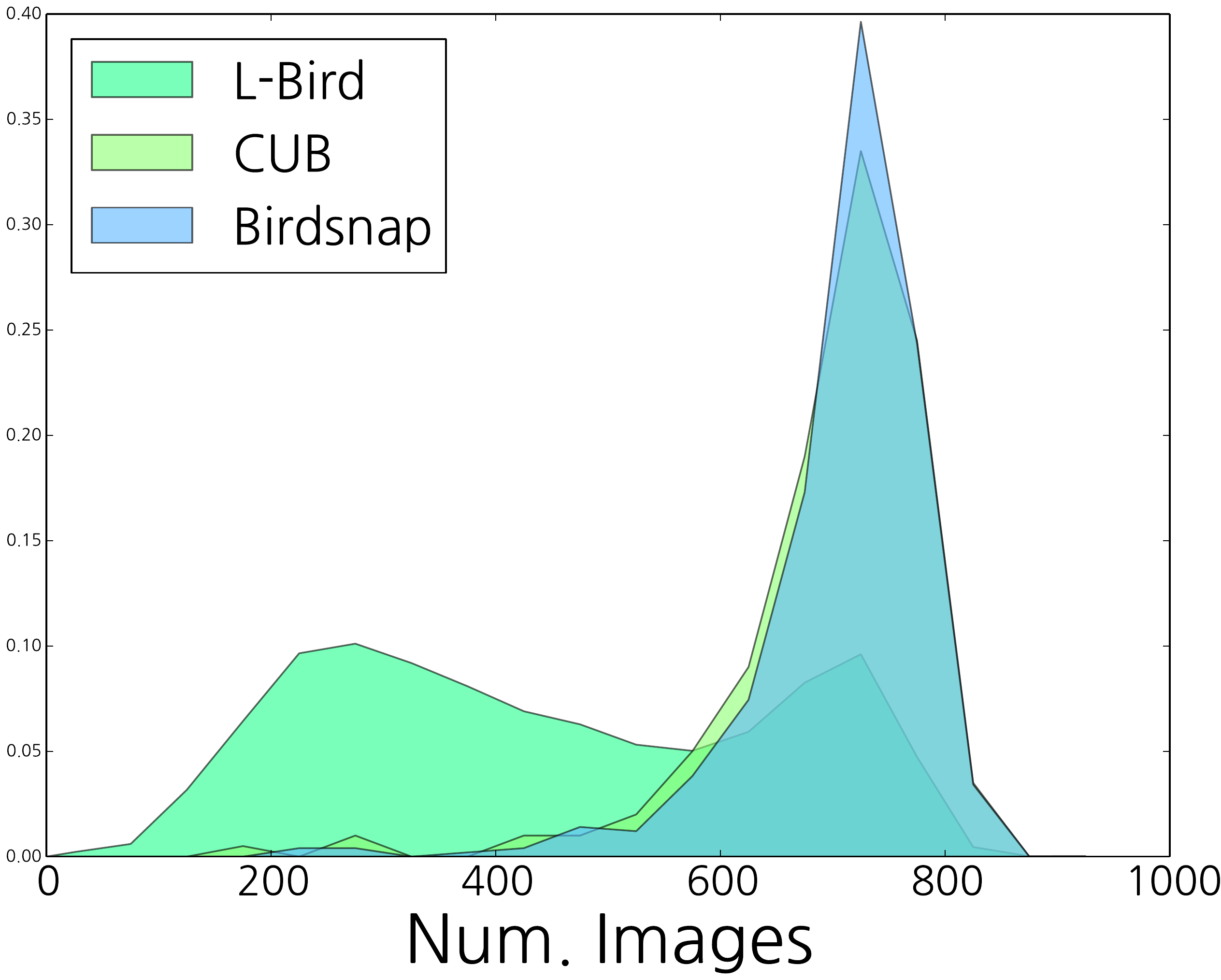}
\includegraphics[width=0.23\linewidth]{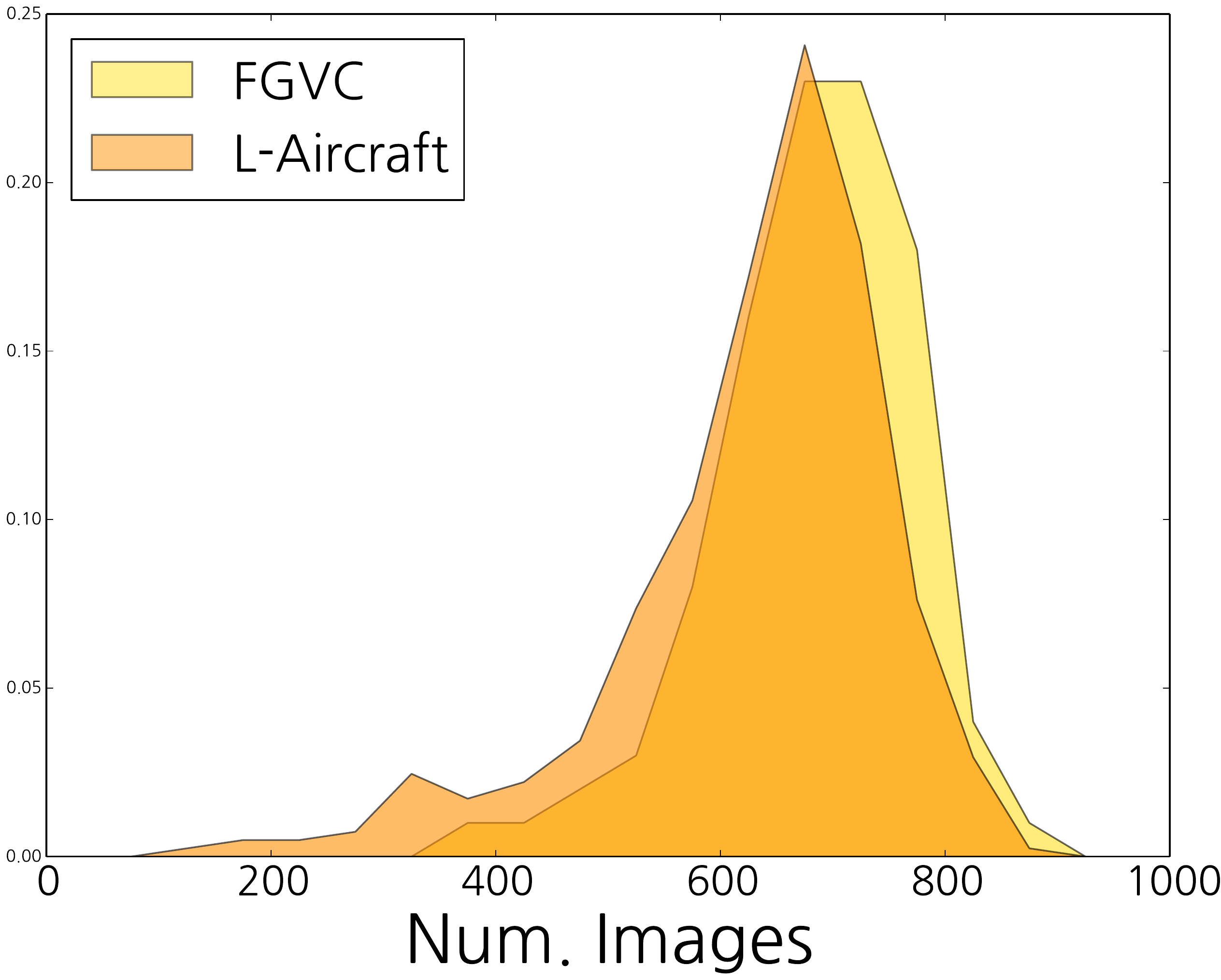}
\includegraphics[width=0.23\linewidth]{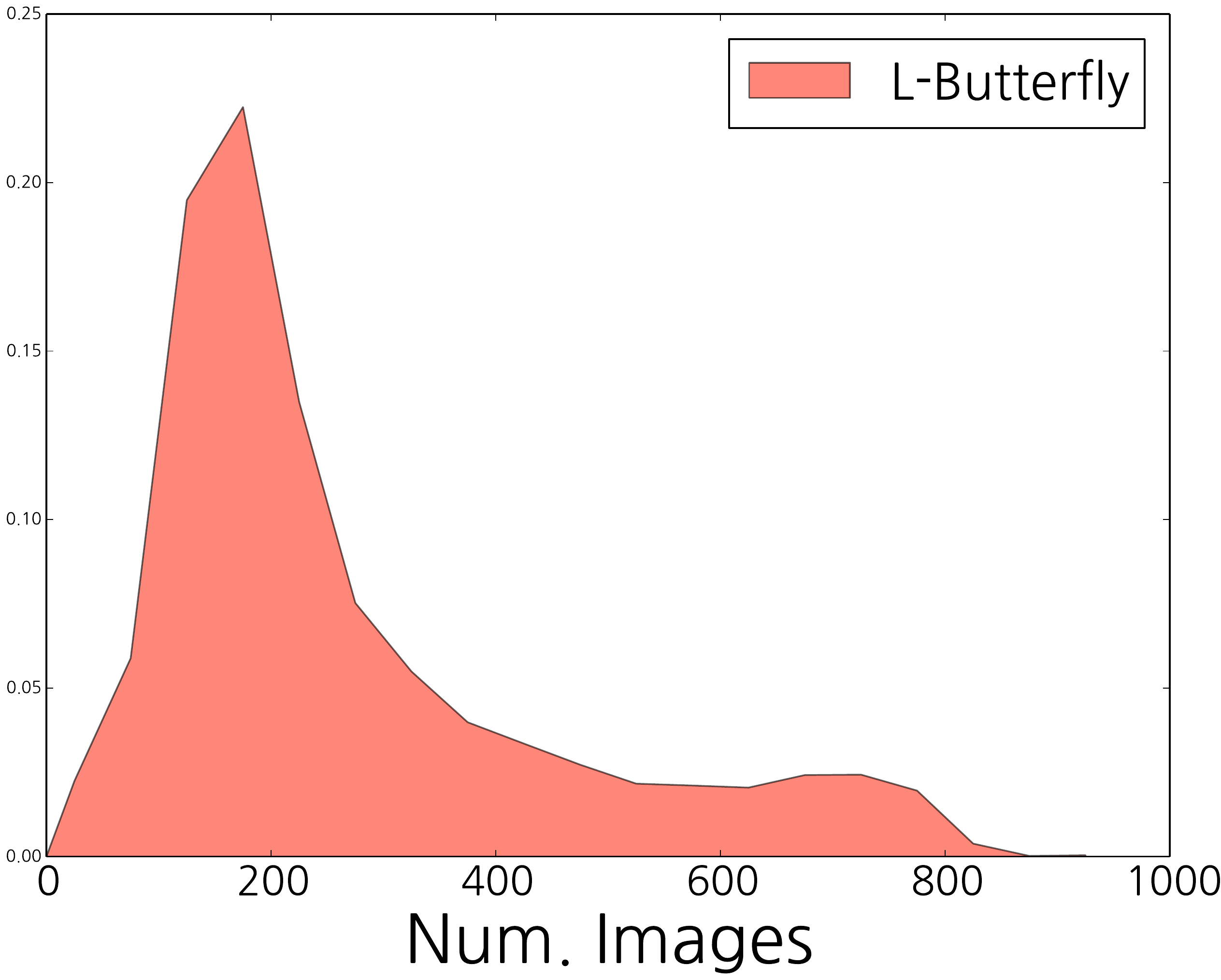}
\includegraphics[width=0.26\linewidth]{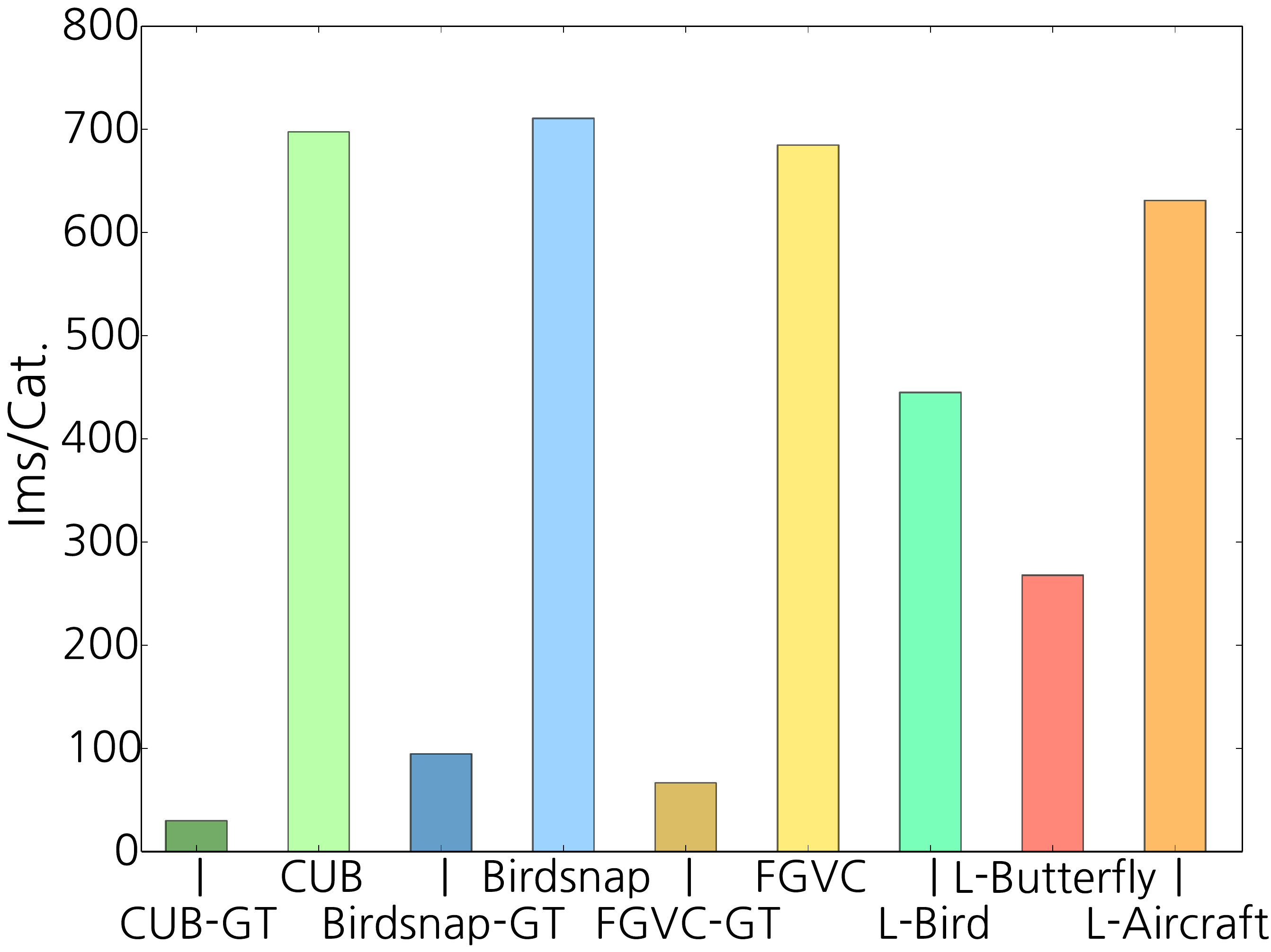}
\caption{
Distributions of the number of images per category available via image search for the categories in CUB, Birdsnap, and \lbird{} (far left), \fgvc{} and \lair{} (middle left), and \llep{} (middle right).
At far right we aggregate and plot the average number of images per category in each dataset in addition to the training sets of each curated dataset we consider, denoted \cubgt{}, \birdsnapgt{}, and \fgvcgt{}.
}
\label{fig:cat_dist}
\end{figure}

\subsubsection{Quantifying the Data.}
How much fine-grained data is available?
In Fig.~\ref{fig:cat_dist} we plot distributions of the number of images retrieved for each category and report aggregates across each set of categories.
We note several trends:
Categories in existing datasets, which are typically common within their fine-grained domain, have more images per category than the long-tail of categories present in the larger \lbird{}, \lair{}, or \llep{}, with the effect most pronounced in \lbird{} and \llep{}.
Further, domains of fine-grained categories have substantially different distributions, \ie \lbird{} and \lair{} have more images per category than \llep{}.
This makes \mbox{sense -- fine-grained} categories and domains of categories that are more common and have a larger enthusiast base will have more imagery since more photos are taken of them.
We also note that results tend to be limited to roughly 800 images per category, even for the most common categories, which is likely a restriction placed on public search results.

Most striking is the large difference between the number of images available via web search and in existing fine-grained datasets: even Birdsnap, which has an average of 94.8 images per category, contains only 13\% as many images as can be obtained with a simple image search.
Though their labels are noisy, web searches unveil an order of magnitude more data which can be used to learn fine-grained categories.

In total, for all four datasets, we obtained 9.8 million images for 26,458 categories, requiring 151.8GB of disk space.\footnote{URLs available at \url{https://github.com/google/goldfinch}}

\begin{figure}[t]
\centering
\includegraphics[width=0.80\linewidth]{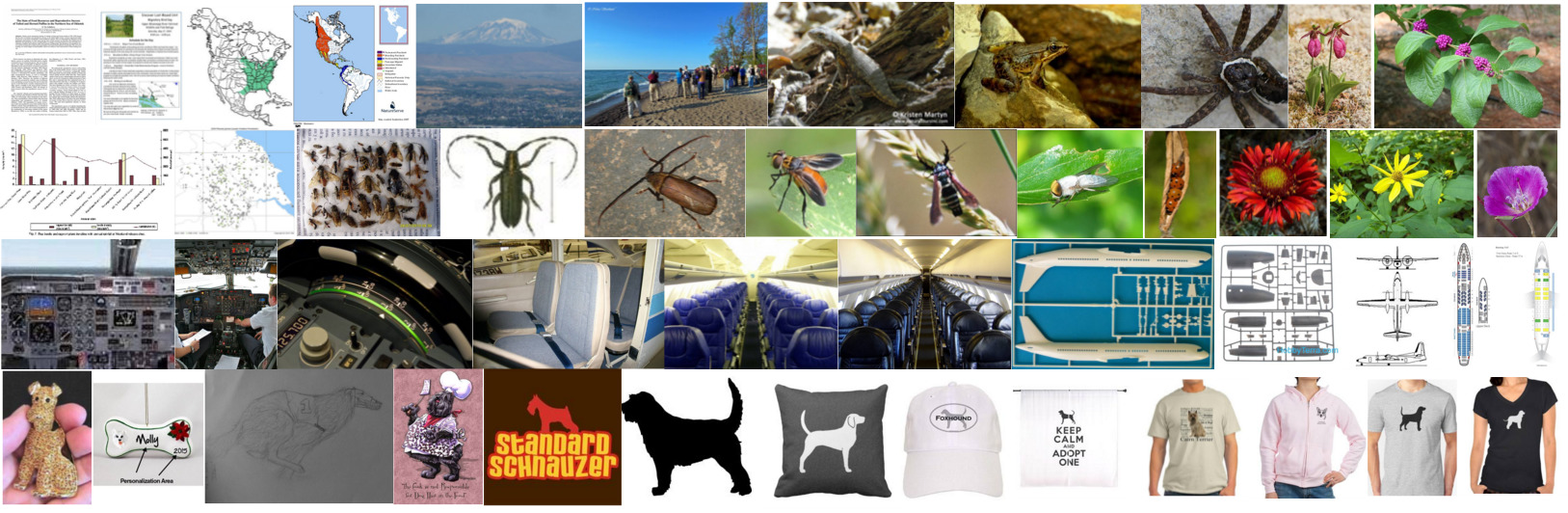}
\caption{
Examples of cross-domain noise for birds, butterflies, airplanes, and dogs.
Images are generally of related categories that are outside the domain of interest, \eg a map of a bird's typical habitat or a t-shirt containing the silhouette of a dog.
}
\label{fig:domain_noise_examples}
\end{figure}

\subsubsection{Noise.}
Though large amounts of imagery are freely available for fine-grained categories, focusing only on scale ignores a key issue: \emph{noise}.
We consider two types of label noise, which we call \emph{cross-domain} noise and \emph{cross-category} noise.
We define cross-domain noise to be the portion of images that are not of any category in the same fine-grained domain, \ie for birds, it is the fraction of images that do not contain a bird (examples in Fig.~\ref{fig:domain_noise_examples}).
In contrast, \emph{cross-category} noise is the portion of images that have the wrong label within a fine-grained domain, \ie an image of a bird with the wrong species label.

\begin{figure}[t]
  \centering
\begin{minipage}{0.40\linewidth}
\centering
\includegraphics[width=0.98\linewidth]{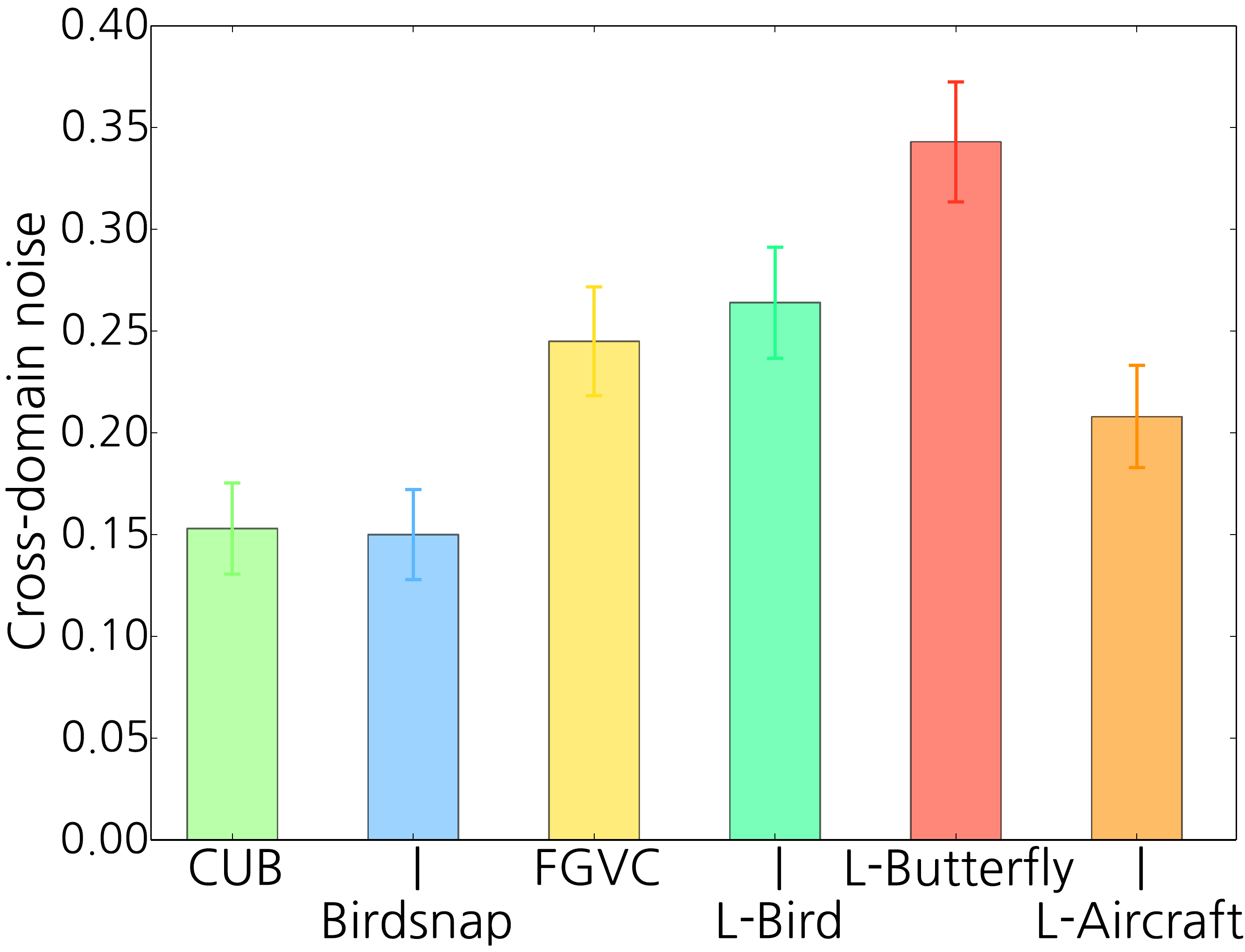}
\caption{The cross-domain noise in search results for each domain.}
\label{fig:cross_domain_stats}
\end{minipage}
\qquad
\begin{minipage}{0.40\linewidth}
\centering
\includegraphics[width=0.98\linewidth]{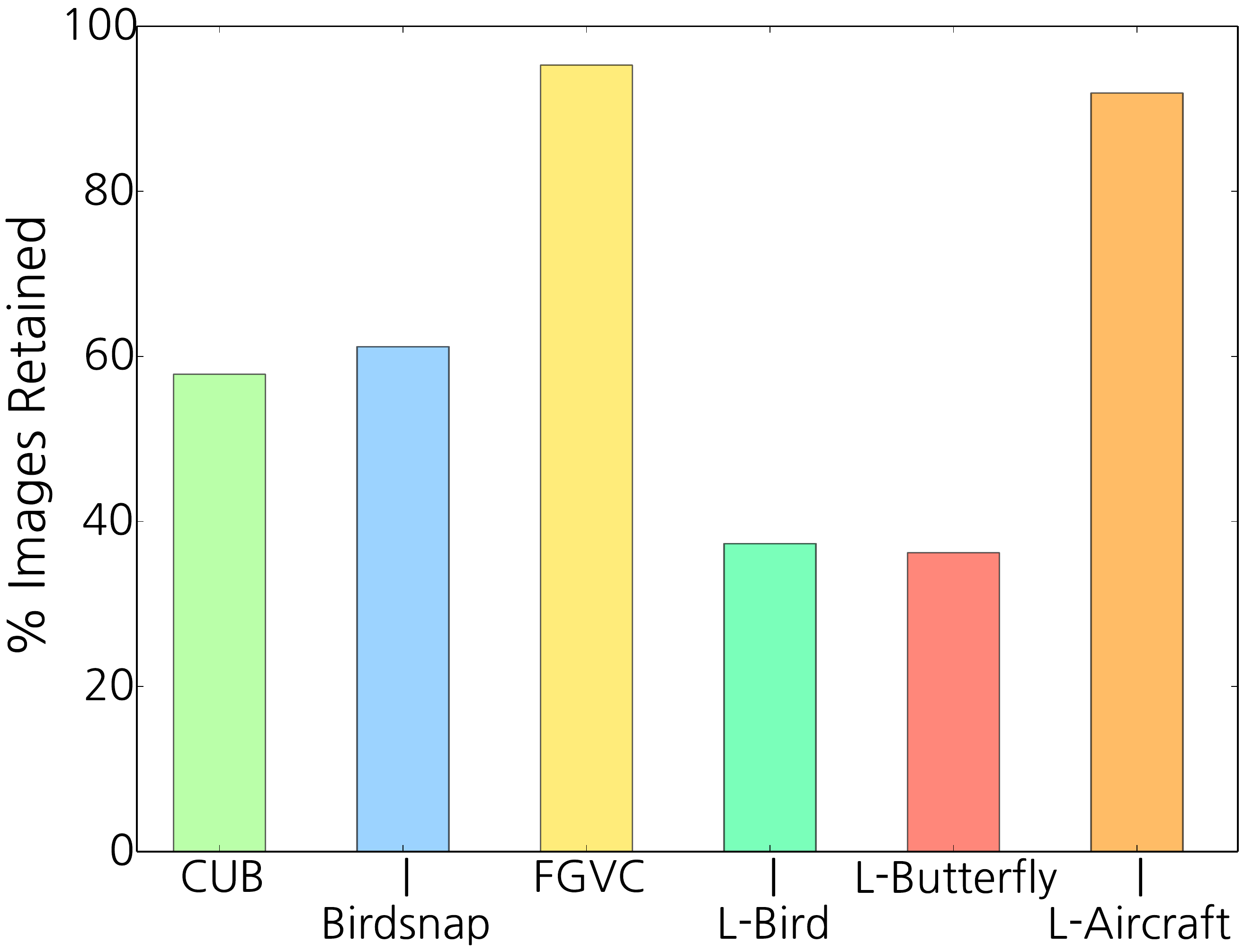}
\caption{The percentage of images retained after filtering.}
\label{fig:filtering_stats}
\end{minipage}
\end{figure}

To quantify levels of cross-domain noise, we manually label a 1,000 image sample from each set of search results, with results in Fig.~\ref{fig:cross_domain_stats}.
Although levels of noise are not too high for any set of categories (max. 34.2\% for \llep{}), we notice an interesting correlation: cross-domain noise decreases moderately as the number of images per category (Fig.~\ref{fig:cat_dist}) increases.
We hypothesize that categories with many search results have a corresponding large pool of images to draw results from, and thus actual search results will tend to be higher-precision.

\begin{figure}[t]
\begin{minipage}[t]{0.48\linewidth}
\centering
\includegraphics[width=0.99\linewidth]{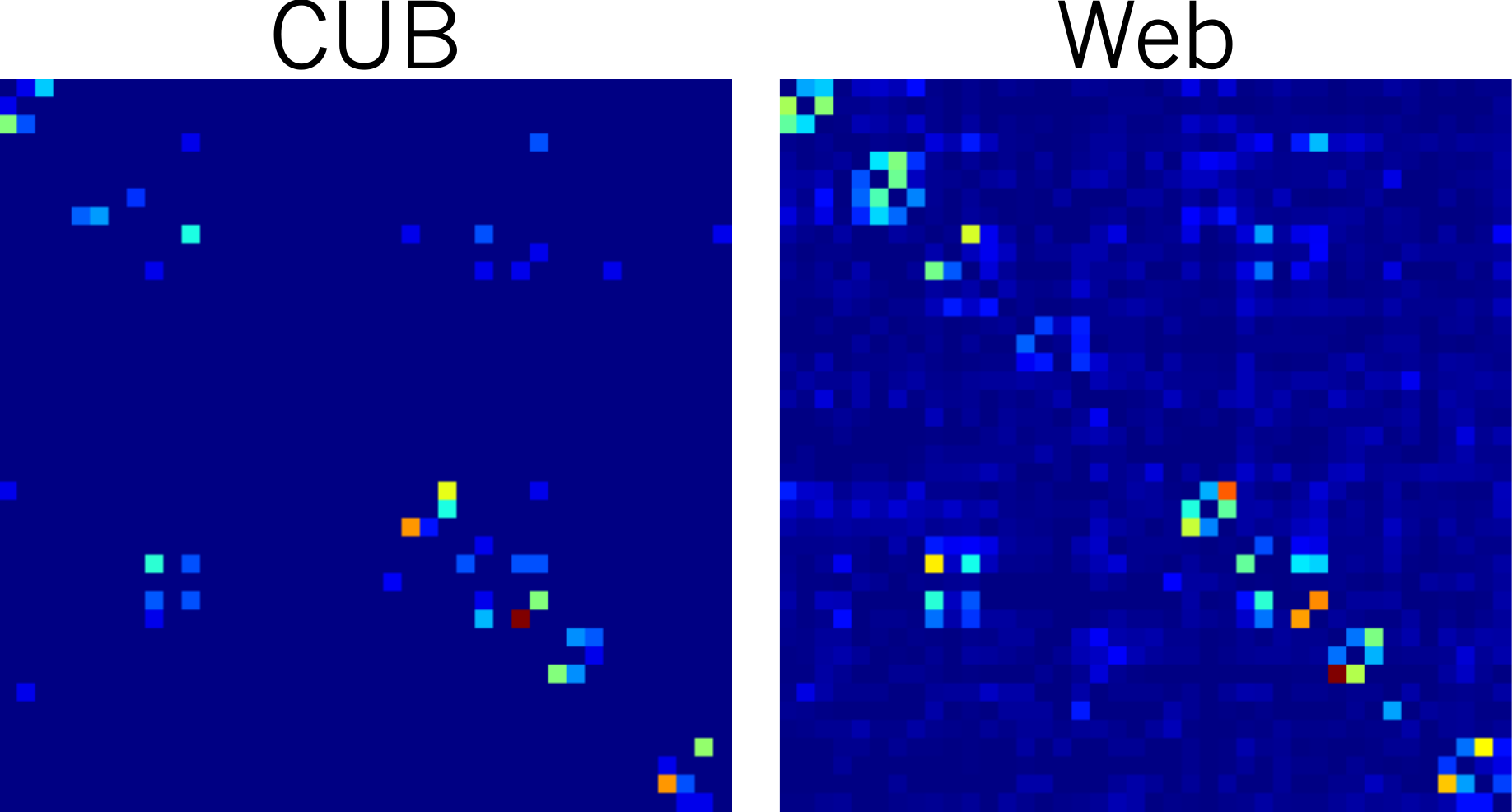}
\caption{\label{fig:conf_mats}
Confusion matrices of the predicted label (column) given the provided label (row) for 30 CUB categories on the CUB test set (left) and search results for CUB categories (right).
For visualization purposes we remove the diagonal.
}
\end{minipage}
\hfill
\begin{minipage}[t]{0.48\linewidth}
\centering
\includegraphics[width=0.99\linewidth]{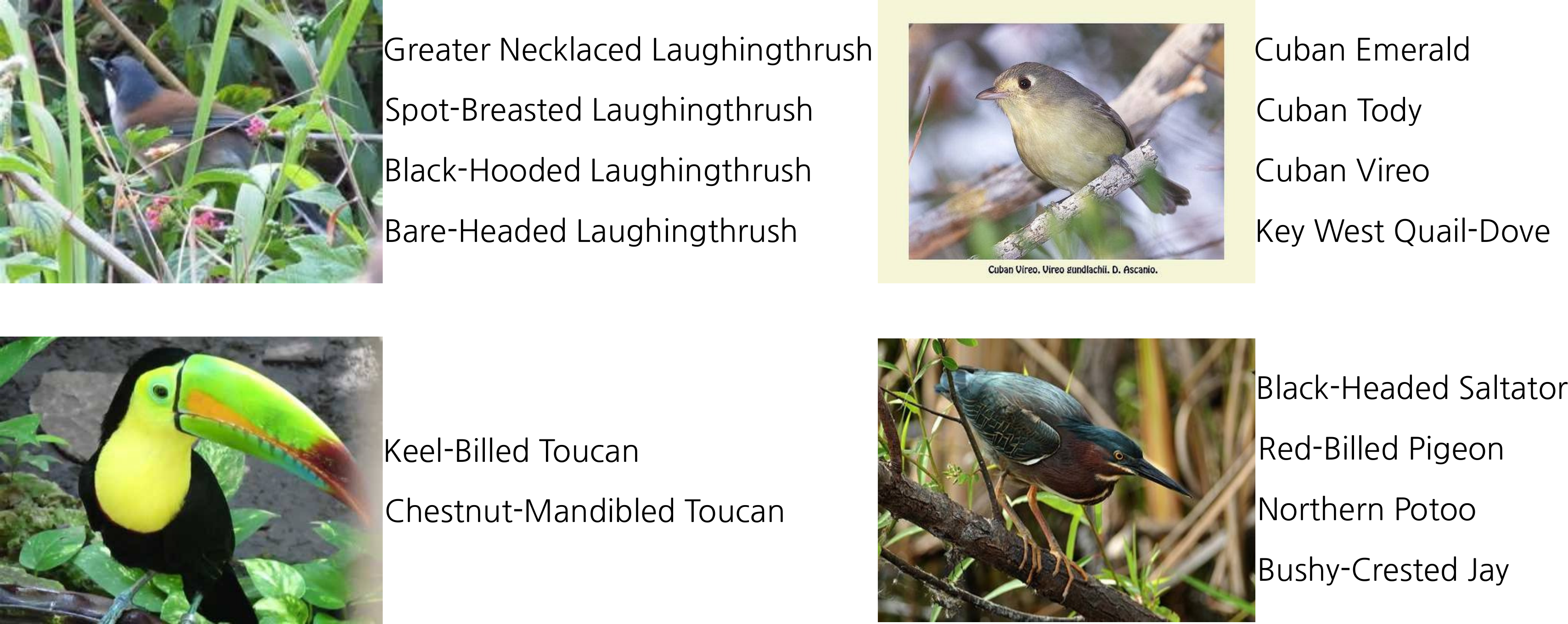}
\caption{\label{fig:filter_example}
Examples of images removed via filtering and the categories whose results they appeared in.
Some share similar names (left examples), while others share similar locations (right examples).
}
\end{minipage}
\end{figure}

In contrast to cross-domain noise, cross-category noise is much harder to quantify, since doing so effectively requires ground truth fine-grained labels of query results.
To examine cross-category noise from at least one vantage point, we show the confusion matrix of given versus predicted labels on 30 categories in the CUB~\cite{wahcub2002011} test set and their web images in Fig.~\ref{fig:conf_mats}, left and right, which we generate via a classifier trained on the CUB training set, acting as a noisy proxy for ground truth labels.
In these confusion matrices, cross-category noise is reflected as a strong off-diagonal pattern, while cross-domain noise would manifest as a diffuse pattern of noise, since images not of the same domain are an equally bad fit to all categories.
Based on this interpretation, the web images show a moderate amount more cross-category noise than the clean CUB test set, though the general confusion pattern is similar.

We propose a simple, yet effective strategy to reduce the effects of cross-category noise: exclude images that appear in search results for more than one category.
This approach, which we refer to as \emph{filtering}, specifically targets images for which there is explicit ambiguity in the category label (examples in Fig.~\ref{fig:filter_example}).
As we demonstrate experimentally, filtering can improve results while reducing training time via the use of a more compact training set -- we show the portion of images kept after filtering in Fig.~\ref{fig:filtering_stats}.
Agreeing with intuition, filtering removes more images when there are more categories.
Anecdotally, we have also tried a few techniques to combat cross-domain noise, but initial experiments did not see any improvement in recognition so we do not expand upon them here.
While reducing cross-domain noise should be beneficial, we believe that it is not as important as cross-category noise in fine-grained recognition due to the absence of out-of-domain classes during testing.

\vspace{-2mm}
\section{Data via Active Learning}
\label{sec:active-learning}
\vspace{-2mm}

In this section we briefly describe an active learning-based approach for collecting large quantities of fine-grained data.
Active learning and other human-in-the-loop systems have previously been used to create datasets in a more cost-efficient way than manual annotation~\cite{yu2015construction,collins2008towards,al-survey}, and our goal is to compare this more traditional approach with simply using noisy data, particularly when considering the application of fine-grained recognition.
In this paper, we apply active learning to the 120 dog breeds in the Stanford Dogs~\cite{khosla2011novel} dataset.

Our system for active learning begins by training a classifier on a seed set of input images and labels (\ie the Stanford Dogs training set), then proceeds by iteratively picking a set of images to annotate, obtaining labels with human annotators, and re-training the classifier.
We use a convolutional neural network~\cite{lecun1998gradient,szegedy2014going,ioffe2015batch} for the classifier, and now describe the key steps of sample selection and human annotation in more detail.

\subsubsection{Sample Selection.}
\begin{figure*}[t]
\centering
\includegraphics[width=.99\linewidth]{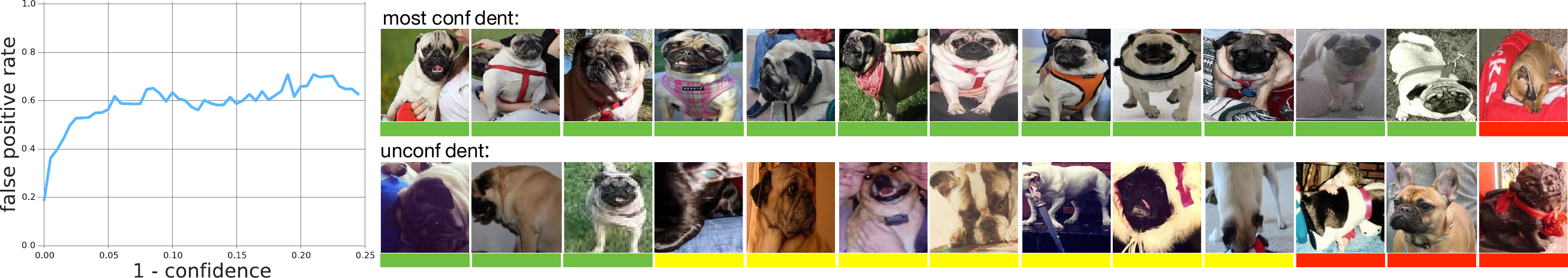}
\caption{
  \textbf{Left:} Classifier confidence versus false positive rate on 100,000 randomly sampled from Flickr images (YFCC100M~\cite{thomee2015yfcc100m}) with dog detections.
  Even the most confident images have a 20\% false positive rate.
\textbf{Right:} Samples from Flickr.  Rectangles below images denote correct ({\color[rgb]{0,0.8,0} green}), incorrect ({\color{red} red}), or ambiguous ({\color[rgb]{0.8,0.8,0.0}yellow}).
\textbf{Top row:} Samples with high confidence for class ``Pug'' from YFCC100M.
\textbf{Bottom row:} Samples with low confidence score for class ``Pug''. 
}
\label{fig:al-fp-rate}
\end{figure*}

There are many possible criterion for sample selection~\cite{al-survey}.
We employ confidence-based sampling: For each category $c$, we select the $b\hat{P}(c)$ images with the top class scores $f_c(x)$ as determined by our current model, where $\hat{P}(c)$ is a desired prior distribution over classes, $b$ is a budget on the number of images to annotate, and $f_c(x)$ is the output of the classifier.
The intuition is as follows: even when $f_c(x)$ is large, false positives still occur quite frequently -- in Fig.~\ref{fig:al-fp-rate} left, observe that the false positive rate is about $20\%$ at the highest confidence range, which might have a large impact on the model.
This contrasts with approaches that focus sampling in uncertain regions~\cite{lewis1994heterogeneous,balcan2007margin,mozafari2014scaling,erkan2010semi}.
We find that images sampled with uncertainty criteria are typically ambiguous and difficult or even impossible for both models \emph{and} humans to annotate correctly, as demonstrated in Fig.~\ref{fig:al-fp-rate} bottom row: unconfident samples are often heavily occluded, at unusual viewpoints, or of mixed, ambiguous breeds, making it unlikely that they can be annotated effectively.
This strategy is similar to the ``expected model change'' sampling criteria~\cite{settles2008multiple}, but done for each class independently.

\subsubsection{Human Annotation.}
\label{sec:raters}
\begin{SCfigure}[][t]
{
\includegraphics[width=0.50\linewidth]{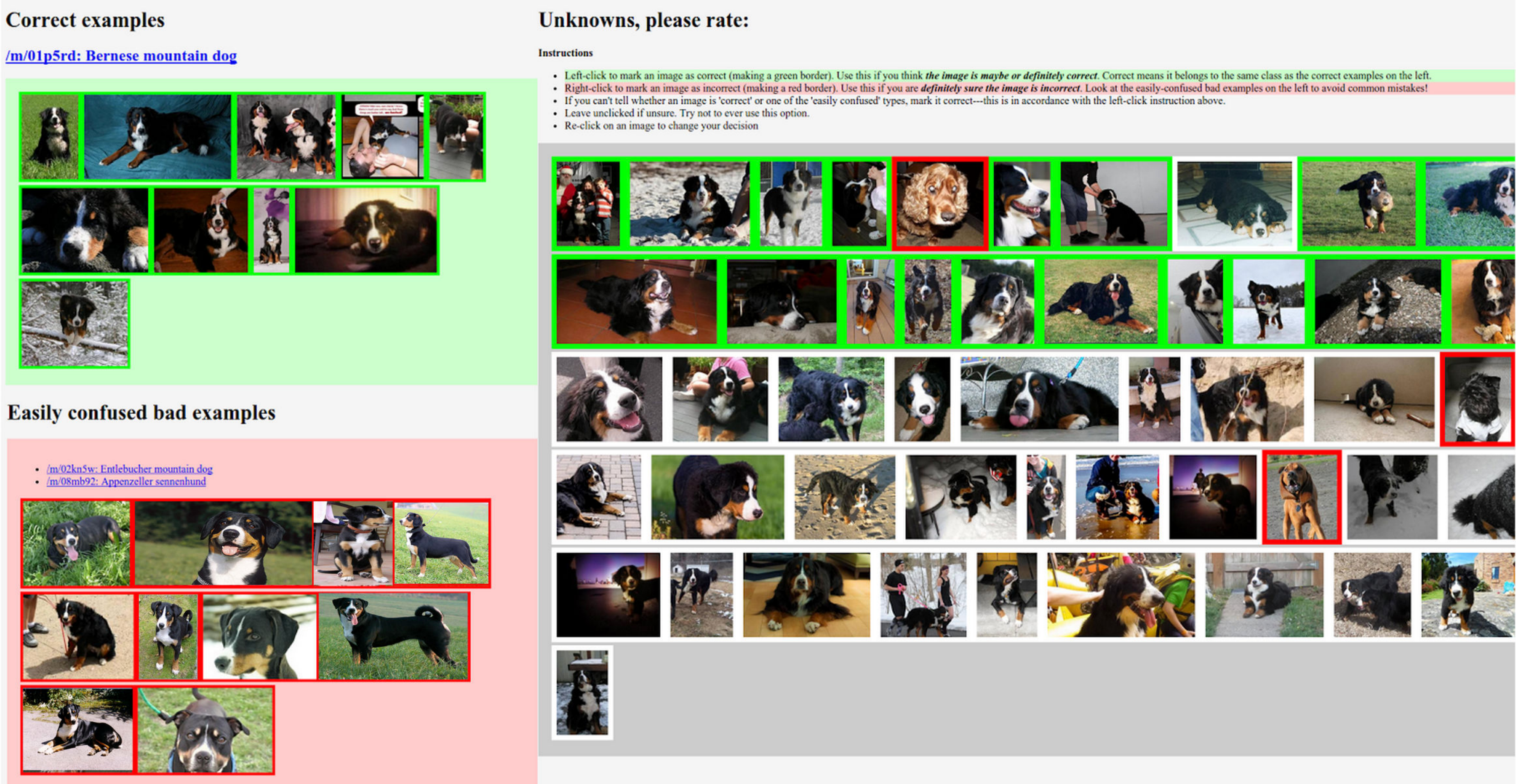}
}
{
\caption{
Our tool for binary annotation of fine-grained categories.
Instructional positive images are provided in the upper left and negatives are provided in the lower left.
}
}
\label{fig:rater-tool}
\end{SCfigure}

Our interface for human annotation of the selected images is shown in Fig.~\ref{fig:rater-tool}.
Careful construction of the interface, including the addition of both positive and negative examples, as well as hidden ``gold standard'' images for immediate feedback, improves annotation accuracy considerably (see Sec.~\ref{app_sec:raters} for quantitative results).
Final category decisions are made via majority vote of three annotators.

\section{Experiments}
\label{sec:experiments}

\subsection{Implementation Details}
The base classifier we use in all noisy data experiments is the Inception-v3 convolutional neural network architecture~\cite{szegedy2015rethinking}, which is among the state of the art methods for generic object recognition~\cite{russakovsky2015ilsvrc,szegedy2016inception,he2015deep}.
Learning rate schedules are determined by performance on a holdout subset of the training data, which is 10\% of the training data for control experiments training on ground truth datasets, or 1\% when training on the larger noisy web data.
Unless otherwise noted, all recognition results use as input a single crop in the center of the image.

Our active learning comparison uses the Yahoo Flickr Creative Commons 100M dataset~\cite{thomee2015yfcc100m} as its pool of unlabeled images, which we first pre-filter with a binary dog classifier and localizer~\cite{szegedy2014going}, resulting in 1.71 million candidate dogs.
We perform up to two rounds of active learning, with a sampling budget $B$ of $10\times$ the original dataset size per round\footnote{To be released.}.
For experiments on Stanford Dogs, we use the CNN of \cite{ioffe2015batch}, which is pre-trained on a version of ILSVRC~\cite{russakovsky2015ilsvrc,imagenet} with dog data removed, since Stanford Dogs is a subset of ILSVRC training data.

\subsection{Removing Ground Truth from Web Images}
One subtle point to be cautious about when using web images is the risk of inadvertently including images from ground truth test sets in the web training data.
To deal with this concern, we performed an aggressive deduplication procedure with all ground truth test sets and their corresponding web images.
This process follows Wang \etal~\cite{wang2014learning}, which is a state of the art method for learning a similarity metric between images.
We tuned this procedure for high near-duplicate recall, manually verifying its quality.
More details are included in the Sec.~\ref{app_sec:dedup}.

\subsection{Main Results}

\begin{table}[t]
\centering
\resizebox{\columnwidth}{!}{%
\begin{tabular}{|l|c|c||l|c|c|}
\hline
Training Data & Acc. & Dataset & Training Data & Acc.& Dataset \\ 
\hhline{|=|=|=||=|=|=|}
\cubgt{} & 84.4 & \multirow{7}{*}{CUB~\cite{wahcub2002011}} &
\fgvcgt{} & 88.1 & \multirow{7}{*}{\fgvc{}~\cite{maji13finegrained}} \\ \cline{1-2} \cline{4-5}
Web (raw) & 87.7 & &
Web (raw) & 90.7 & \\ \cline{1-2} \cline{4-5}
Web (filtered) & 89.0 & &
Web (filtered) & 91.1 & \\ \cline{1-2} \cline{4-5}
\lbird{} & 91.9 & &
\lair{} & 90.9 & \\ \cline{1-2} \cline{4-5}
\lbird{}(MC) & 92.3 & &
\lair{}(MC) & 93.4 & \\ \cline{1-2} \cline{4-5}
\lbird{}+\cubgt{} & 92.2 & &
\lair{}+\fgvcgt{} & 94.5 & \\ \cline{1-2} \cline{4-5}
\lbird{}+\cubgt{}(MC) & 92.8 & &
\lair{}+\fgvcgt{}(MC) & 95.9 & \\ 
\hhline{|=|=|=#=|=|=|}
\birdsnapgt{} & 78.2 & \multirow{7}{*}{Birdsnap~\cite{bergbirdsnapcvpr2014}} &
Stanford-GT & 80.6 & \multirow{7}{*}{Stanford Dogs~\cite{khosla2011novel}} \\ \cline{1-2} \cline{4-5}
Web (raw) & 76.1 & &
Web (raw) & 78.5 & \\ \cline{1-2} \cline{4-5}
Web (filtered) & 78.2 & &
Web (filtered) & 78.4 & \\ \cline{1-2} \cline{4-5}
\lbird{} & 82.8 & &
L-Dog & 78.4 & \\ \cline{1-2} \cline{4-5}
\lbird{}(MC) & 85.4 & &
L-Dog(MC) & 80.8 & \\ \cline{1-2} \cline{4-5}
\lbird{}+\birdsnapgt{} & 83.9 & &
L-Dog+Stanford-GT & 84.0 & \\ \cline{1-2} \cline{4-5}
\lbird{}+\birdsnapgt{}(MC) & 85.4 & &
L-Dog+Stanford-GT(MC) & 85.9 & \\ 
\hline
\end{tabular}
}
\caption{
Comparison of data source used during training with recognition performance, given in terms of Top-1 accuracy.
``\cubgt{}'' indicates training only on the ground truth CUB training set, ``Web (raw)'' trains on all search results for CUB categories, and ``Web (filtered)'' applies filtering between categories within a domain (birds).
\lbird{} denotes training first on \lbird{}, then fine-tuning on the subset of categories under evaluation (\ie the filtered web images), and \lbird{}+\cubgt{} indicates training on \lbird{}, then fine-tuning on Web (filtered), and finally fine-tuning again on \cubgt{}.
Similar notation is used for the other datasets.
\mbox{``(MC)''} indicates using multiple crops at test time (see text for details).
We note that only the rows with \mbox{``-GT''} make use of the ground truth training set; all other rows rely solely on noisy web imagery.
}
\label{table:rec_results}
\end{table}

We present our main recognition results in Tab.~\ref{table:rec_results}, where we compare performance when the training set consists of either the ground truth training set, raw web images of the categories in the corresponding evaluation dataset, web images after applying our filtering strategy, all web images of a particular domain, or all images including even the ground truth training set.

On CUB-200-2011~\cite{wahcub2002011}, the smallest dataset we consider, even using raw search results as training data results in a better model than the annotated training set, with filtering further improving results by 1.3\%.
For Birdsnap~\cite{bergbirdsnapcvpr2014}, the largest of the ground truth datasets we evaluate on, raw data mildly underperforms using the ground truth training set, though filtering improves results to be on par.
On both CUB and Birdsnap, training first on the very large set of categories in \lbird{} results in dramatic improvements, improving performance on CUB further by 2.9\% and on Birdsnap by 4.6\%.
This is an important point: even if the end task consists of classifying only a small number of categories, training with more fine-grained categories yields significantly more effective networks.
This can also be thought of as a form of transfer learning within the same fine-grained domain, allowing features learned on a related task to be useful for the final classification problem.
When permitted access to the annotated ground truth training sets for additional fine-tuning and domain transfer, results increase by another $0.3\%$ on CUB and $1.1\%$ on Birdsnap.

For the aircraft categories in \fgvc{}, results are largely similar but weaker in magnitude.
Training on raw web data results in a significant gain of 2.6\% compared to using the curated training set, and filtering, which did not affect the size of the training set much (Fig.~\ref{fig:filtering_stats}), changes results only slightly in a positive direction.
Counterintuitively, pre-training on a larger set of aircraft does not improve results on \fgvc{}.
Our hypothesis for the difference between birds and aircraft in this regard is this: since there are many more species of birds in \lbird{} than there are aircraft in \lair{} (10,982 vs 409), not only is the training size of \lbird{} larger\no{ (1.8M images vs 236k)}, but each training example provides stronger information because it distinguishes between a larger set of mutually-exclusive categories.
Nonetheless, when access to the curated training set is available for fine-tuning, performance dramatically increases to 94.5\%.
On Stanford Dogs we see results similar to \fgvc{}, though for dogs we happen to see a mild loss when comparing to the ground truth training set, not much difference with filtering or using \ldog{}, and a large boost from adding in the ground truth training set.

An additional factor that can influence performance of web models is domain shift -- if images in the ground truth test set have very different visual properties compared to web images, performance will naturally differ.
Similarly, if category names or definitions within a dataset are even mildly off, web-based methods will be at a disadvantage without access to the ground truth training set.
Adding the ground truth training data fixes this domain shift, making web-trained models quickly recover, with a particularly large gain if the network has already learned a good representation, matching the pattern of results for Stanford Dogs.

\subsubsection{Limits of Web-Trained Models.}
To push our models to their limits, we additionally evaluate using 144 image crops at test time, averaging predictions across each crop, denoted ``(MC)'' in Tab.~\ref{table:rec_results}.
This brings results up to 92.3\%/92.8\% on CUB (without/with CUB training data), 85.4\%/85.4\% on Birdsnap, 93.4\%/95.9\% on \fgvc{}, and 80.8\%/85.9\% on Stanford Dogs.
We note that this is close to human expert performance on CUB, which is estimated to be between $93\%$~\cite{branson2014ignorant} and $95.6\%$~\cite{horn2015}.

\begin{SCtable}[][t]
{
\resizebox{0.6\columnwidth}{!}{%
\begin{tabular}{|l|l|c|}
\hline
Method & Training Annotations & Acc. \\ \hhline{|=|=|=|}
Alignments~\cite{gavves2014local}& GT & 53.6 \\ \hline
PDD~\cite{simon14pdd} & GT+BB+Parts & 60.6 \\ \hline
PB R-CNN~\cite{zhang2014part} & GT+BB+Parts & 73.9 \\ \hline
Weak Sup.~\cite{zhang2015weakly} & GT & 75.0 \\ \hline
PN-DCN~\cite{branson2014bird} & GT+BB+Parts & 75.7 \\ \hline
Two-Level~\cite{xiao2015application} & GT &  77.9 \\ \hline
Consensus~\cite{shih2015part} & GT+BB+Parts &  78.3 \\ \hline
NAC~\cite{simon2015neural} & GT & 81.0 \\ \hline
FG-Without~\cite{krause2015fine} & GT+BB & 82.0 \\ \hline
STN~\cite{jaderberg2015spatial} & GT & 84.1 \\ \hline
Bilinear~\cite{lin2015bilinear} & GT & 84.1 \\ \hline
Augmenting~\cite{xu2015augmenting} & GT+BB+Parts+Web & 84.6 \\ \hline
\bf{Noisy Data+CNN~\textnormal{\cite{szegedy2015rethinking}}} & \bf{Web} & \bf{92.3} \\ \hline
\end{tabular}
}
}
{\caption{Comparison with prior work on CUB-200-2011~\cite{wahcub2002011}.
We only include methods which use no annotations at test time.
Here ``GT'' refers to using Ground Truth category labels in the training set of CUB, ``BBox'' indicates using bounding boxes, and ``Parts'' additionally uses part annotations.
}}
\label{table:cub_prior}
\end{SCtable}

\subsubsection{Comparison with Prior Work.}
We compare our results to prior work on CUB, the most competitive fine-grained dataset, in Tab.~\ref{table:cub_prior}.
While even our baseline model using only ground truth data from Tab.~\ref{table:rec_results} was at state of the art levels, by forgoing the CUB training set and only training using noisy data from the web, our models greatly outperform all prior work.
On \fgvc{}, which is more recent and fewer works have evaluated on, the best prior performing method we are aware of is the Bilinear CNN model of Lin \etal~\cite{lin2015bilinear}, which has accuracy 84.1\% (ours is 93.4\% without \fgvc{} training data, 95.9\% with), and on Birdsnap, which is even more recent, the best performing method we are aware of that uses no extra annotations during test time is the original 66.6\% by Berg \etal~\cite{bergbirdsnapcvpr2014} (ours is 85.4\%).
On Stanford Dogs, the most competitive related work is~\cite{sermanet2014attention}, which uses an attention-based recurrent neural network to achieve $76.8\%$ (ours is $80.8\%$ without ground truth training data, $85.9\%$ with).

We identify two key reasons for these large improvements:
The first is the use of a strong generic classifier~\cite{szegedy2015rethinking}.
A number of prior works have identified the importance of having well-trained CNNs as components in their systems for fine-grained recognition~\cite{lin2015bilinear,jaderberg2015spatial,krause2015fine,zhang2014part,branson2014bird}, which our work provides strong evidence for.
On all four evaluation datasets, our CNN of choice~\cite{szegedy2015rethinking}, trained on the ground truth training set alone and without any architectural modifications, performs at levels at or above the previous state-of-the-art.
The second reason for improvement is the large utility of noisy web data for fine-grained recognition, which is the focus of this work.

We finally remind the reader that our work focuses on the application-level problem of recognizing a given set of fine-grained categories, which might not come with their own expert-annotated training images.
The use of existing test sets serves to provide an accurate measure of performance and put our work in a larger context, but results may not be strictly comparable with prior work that operates within a single given dataset.


\subsubsection{Comparison with Active Learning.}
We compare using noisy web data with a more traditional active learning-based approach (Sec.~\ref{sec:active-learning}) under several different settings in Tab.~\ref{tab:al-results}.
We first verify the efficacy of active learning itself: when training the network from scratch (\ie no fine-tuning), active learning improves performance by up to $15.6\%$, and when fine-tuning, results still improve by $1.5\%$.

How does active learning compare to using web data?
Purely using filtered web data compares favorably to non-fine-tuned active learning methods ($4.4\%$ better), though lags behind the fine-tuned models somewhat.
To better compare the active learning and noisy web data, we factor out the difference in scale by performing an experiment with subsampled active learning data, setting it to be the same size as the filtered web data.
Surprisingly, performance is very similar, with only a $0.4\%$ advantage for the cleaner, annotated active learning data, highlighting the effectiveness of noisy web data despite the lack of manual annotation.
If we furthermore augment the filtered web images with the Stanford Dogs training set, which the active learning method notably used both as training data and its seed set of images, performance improves to even be slightly better than the manually-annotated active learning data ($0.5\%$ improvement).

\begin{SCtable}[][t]
\centering
\begin{tabular}{|l|c|r|}
\hline
Training Procedure & Acc. \\
\hhline{|=|=|}
Stanford-GT (scratch)                      & 58.4 \\ \hline
A.L., one round (scratch)                        & 65.8 \\ \hline
A.L., two rounds (scratch)                        & 74.0 \\ \hline
Stanford-GT (ft)                               & 80.6 \\ \hline
A.L., one round (ft)                          & 81.6 \\ \hline
A.L., one round (ft, subsample)                          & 78.8 \\ \hline
A.L., two rounds (ft)                          & 82.1 \\ \hline
Web (filtered)                                   & 78.4 \\ \hline
Web (filtered) + Stanford-GT                           & 82.6  \\
\hline
\end{tabular}
\caption{
Active learning-based results on Stanford Dogs~\cite{khosla2011novel}, presented in terms of top-1 accuracy.
Methods with ``(scratch)'' indicate training from scratch and ``(ft)'' indicates fine-tuning from a network pre-trained on ILSVRC, with web models also fine-tuned.
``subsample'' refers to downsampling the active learning data to be the same size as the filtered web images.
Note that Stanford-GT is a subset of active learning data, which is denoted ``A.L.''.
}
\label{tab:al-results}
\end{SCtable}

These experiments indicate that, while more traditional active learning-based approaches towards expanding datasets are effective ways to improve recognition performance given a suitable budget, simply using noisy images retrieved from the web can be nearly as good, if not better.
As web images require no manual annotation and are openly available, we believe this is strong evidence for their use in solving fine-grained recognition.

\subsubsection{Very Large-Scale Fine-Grained Recognition.}

A key advantage of using noisy data is the ability to scale to large numbers of fine-grained classes.
However, this poses a challenge for evaluation -- it is infeasible to manually annotate images with one of the 10,982 categories in \lbird{}, 14,553 categories in \llep{}, and would even be very time-consuming to annotate images with the 409 categories in \lair{}.
Therefore, we turn to an approximate evaluation, establishing a rough estimate on true performance.
Specifically, we query Flickr for up to 25 images of each category, keeping only those images whose title strictly contains the name of each category, and aggressively deduplicate these images with our training set in order to ensure a fair evaluation.
Although this is not a perfect evaluation set, and is thus an area where annotation of fine-grained datasets is particularly valuable~\cite{horn2015}, we find that it is remarkably clean on the surface:
based on a 1,000-image estimate, we measure the cross-domain noise of \lbird{} at only 1\%, \llep{} at 2.3\%, and \lair{} at 4.5\%.
An independent evaluation~\cite{horn2015} further measures all sources of noise combined to be only 16\% when searching for bird species.
In total, this yields 42,115 testing images for \lbird{}, 42,046 for \llep{}, and 3,131 for \lair{}.

\begin{figure*}[t]
\centering
\includegraphics[width=0.99\linewidth]{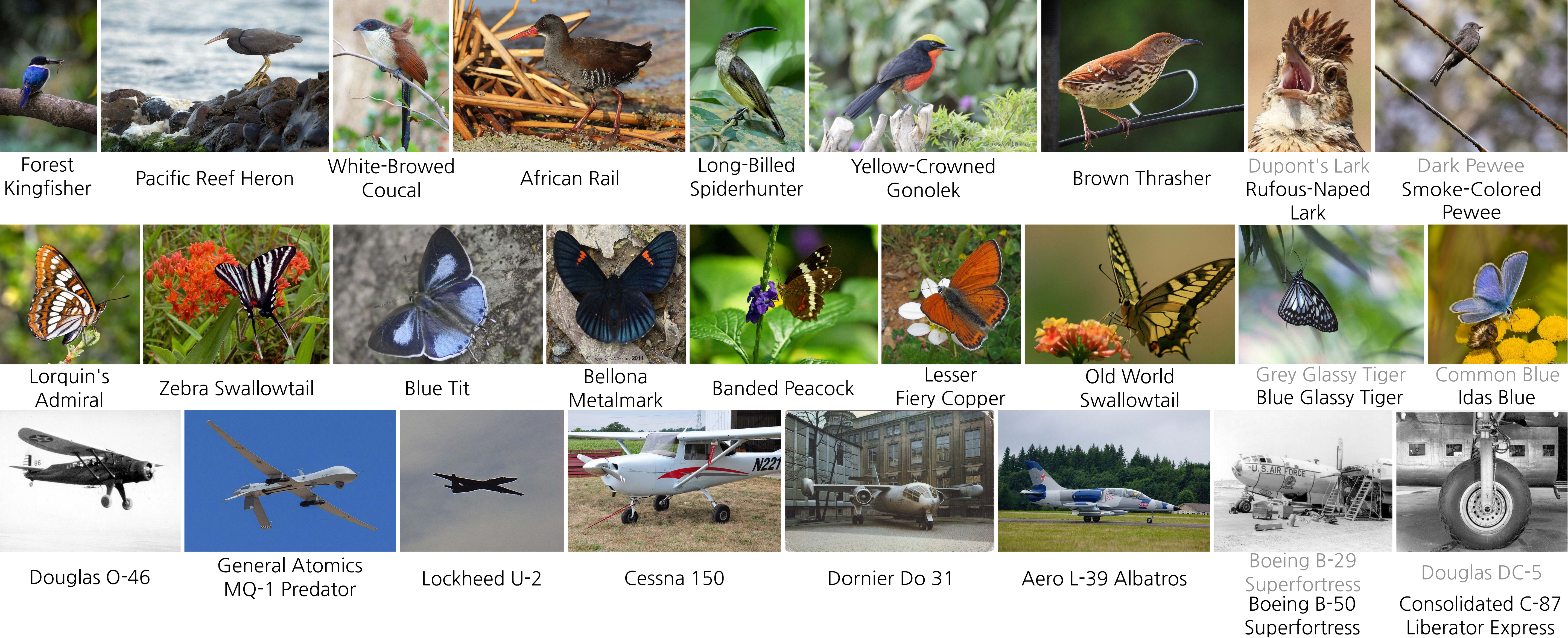}
\caption{
Classification results on very large-scale fine-grained recognition.
From top to bottom, depicted are examples of categories in \lbird{}, \llep{}, and \lair{}, along with their category name.
The first examples in each row are correctly predicted by our models, while the last two examples in each row are errors, with our prediction in grey and correct category (according to Flickr metadata) printed below.
}
\label{fig:flickr_qualitative}
\end{figure*}

Given the difficulty and noise, performance is surprisingly high:
On \lbird{} top-1 accuracy is 73.1\%/75.8\% (1/144 crops), for \llep{} it is 65.9\%/68.1\%, and for \lair{} it is 72.7\%/77.5\%.
Corresponding mAP numbers, which are better suited for handling class imbalance, are 61.9, 54.8, and 70.5, reported for the single crop setting.
We show qualitative results in Fig.~\ref{fig:flickr_qualitative}.
These categories span multiple continents in space (birds, butterflies) and decades in time (aircraft), demonstrating the breadth of categories in the world that can be recognized using only public sources of noisy fine-grained data.
To the best of our knowledge, these results represent the largest number of fine-grained categories distinguished by any single system to date.

\subsubsection{How Much Data is Really Necessary?}
In order to better understand the utility of noisy web data for fine-grained recognition, we perform a control experiment on the web data for CUB.
Using the filtered web images as a base, we train models using progressively larger subsets of the results as training data, taking the top ranked images across categories for each experiment.
Performance versus the amount of training data is shown in Fig.~\ref{fig:cub_data_acc}.
Surprisingly, relatively few web images are required to do as well as training on the CUB training set, and adding more noisy web images always helps, even when at the limit of search results.
Based on this analysis, we estimate that one noisy web image for CUB categories is ``worth'' 0.507 ground truth training images~\cite{torralba2011unbiased}.

\subsubsection{Error Analysis.}

\begin{figure}[t]
\begin{minipage}[t]{0.49\linewidth}
\centering
\includegraphics[width=0.99\linewidth]{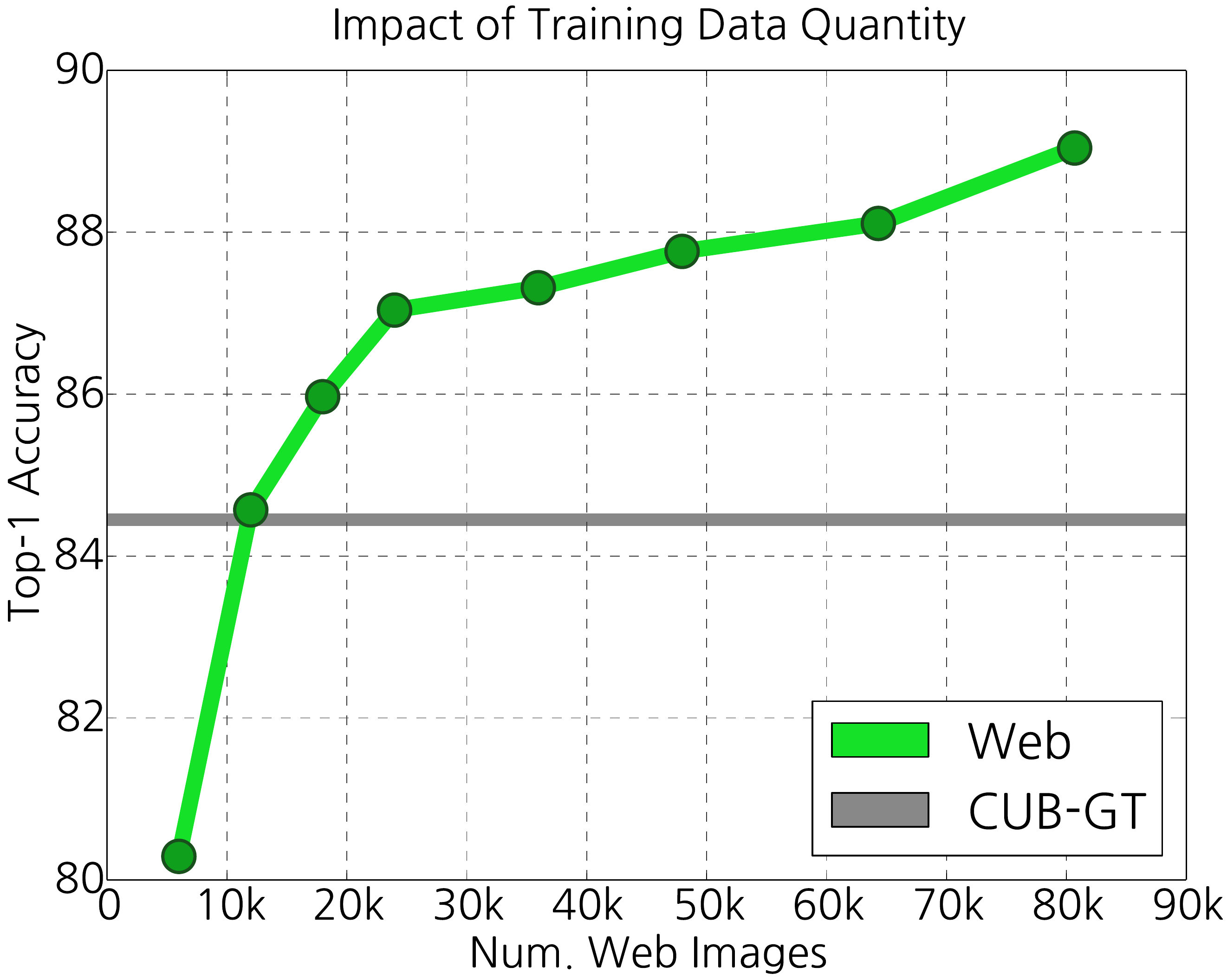}
\caption{Number of web images used for training vs. performance on CUB-200-2011~\cite{wahcub2002011}.
  We vary the amount of web training data in multiples of the CUB training set size (5,994 images).
  Also shown is performance when training on the ground truth CUB training set (CUB-GT).
}
\label{fig:cub_data_acc}
\end{minipage}
\hfill
\begin{minipage}[t]{0.49\linewidth}
\centering
\includegraphics[width=0.99\linewidth]{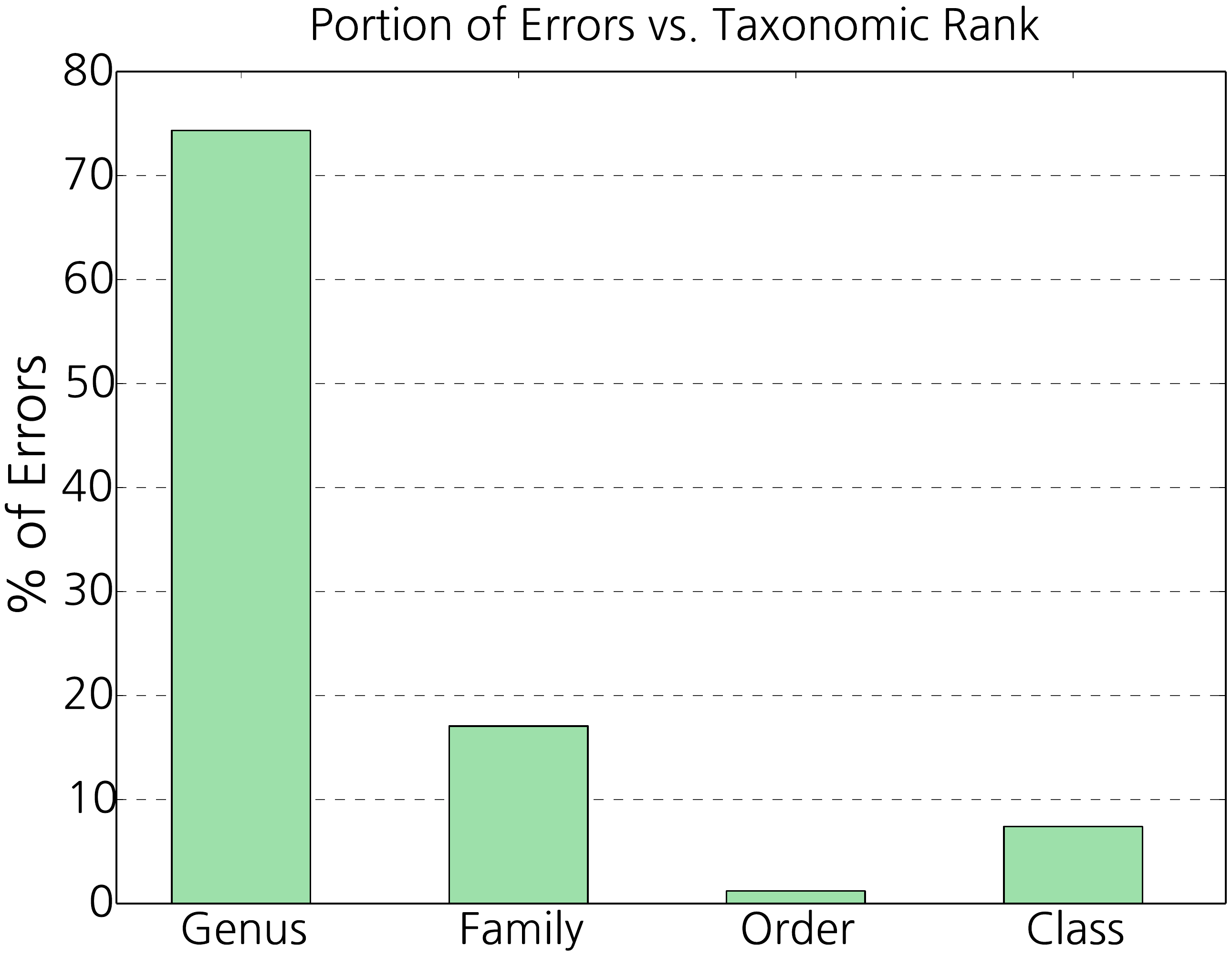}
\caption{
The errors on \lbird{} that fall in each taxonomic rank, represented as a portion of all errors made.
For each error made, we calculate the taxonomic rank of the least common ancestor of the predicted and test category.
}
\label{fig:hierarchy_confusion}
\end{minipage}
\end{figure}

Given the high performance of these models, what room is left for improvement?
In Fig.~\ref{fig:hierarchy_confusion} we show the taxonomic distribution of the remaining errors on \lbird{}.
The vast majority of errors (74.3\%) are made between very similar classes at the genus level, indicating that most of the remaining errors are indeed between extremely similar categories, and only very few errors (7.4\%) are made between dissimilar classes, whose least common ancestor is the ``Aves'' (\ie Bird) taxonomic class.
This suggests that most errors still made by the models are fairly reasonable, corroborating the qualitative results of Fig.~\ref{fig:flickr_qualitative}.

\section{Discussion}
\label{sec:conclusion}
In this work we have demonstrated the utility of noisy data toward solving the problem of fine-grained recognition.
We found that the combination of a generic classification model and web data, filtered with a simple strategy, was surprisingly effective at discriminating fine-grained categories.
This approach performs favorably when compared to a more traditional active learning method for expanding datasets, but is even more scalable, which we demonstrated experimentally on up to 14,553 fine-grained categories.
One potential limitation of the approach is the availability of imagery for categories either not found or not described in the public domain, for which an alternative method such as active learning may be better suited.
Another limitation is the current focus on classification, which may be problematic if applications arise where multiple objects are present or localization is otherwise required.
Nonetheless, with these insights on the unreasonable effectiveness of noisy data, we are optimistic for applications of fine-grained recognition in the near future.

\section{Acknowledgments}
\label{sec:ack}
We thank Gal Chechik, Chuck Rosenberg, Zhen Li, Timnit Gebru, Vignesh Ramanathan, Oliver Groth, and the anonymous reviewers for valuable feedback.

\clearpage


\appendix
\noindent
{\LARGE{\textbf{Appendix}}}

\section{Active Learning Details}
Here we provide additional details for our active learning baseline, including further description of the interface, improvements in rater quality as a result of this interface, statistics of the number of positives obtained per class in each round of active learning, and qualitative examples of images obtained.

\subsection{Interface}

Designing an effective rater tool is of critical importance when getting non-experts to rate fine-grained categories.
We seek to give the raters simple decisions and to provide them with as much information as possible to make the correct decision in a generic and scalable way.
Fig.~\ref{fig:rater-tool2} shows our rater interface, which includes the following components to serve this purpose:

\subsubsection{Instructional positive images} inform the rater of within-class variation.
These images are obtained from the seed dataset input to active learning.
Many rater tools only provide this (\eg\cite{mscoco}), which does not provide a clear class boundary concept on its own.
We also provide links to Google Image Search and encourage raters to research the full space of examples of the class concept.  

\subsubsection{Instructional negative images} 
\begin{figure*}[t]
\centering
\includegraphics[width=0.85\linewidth]{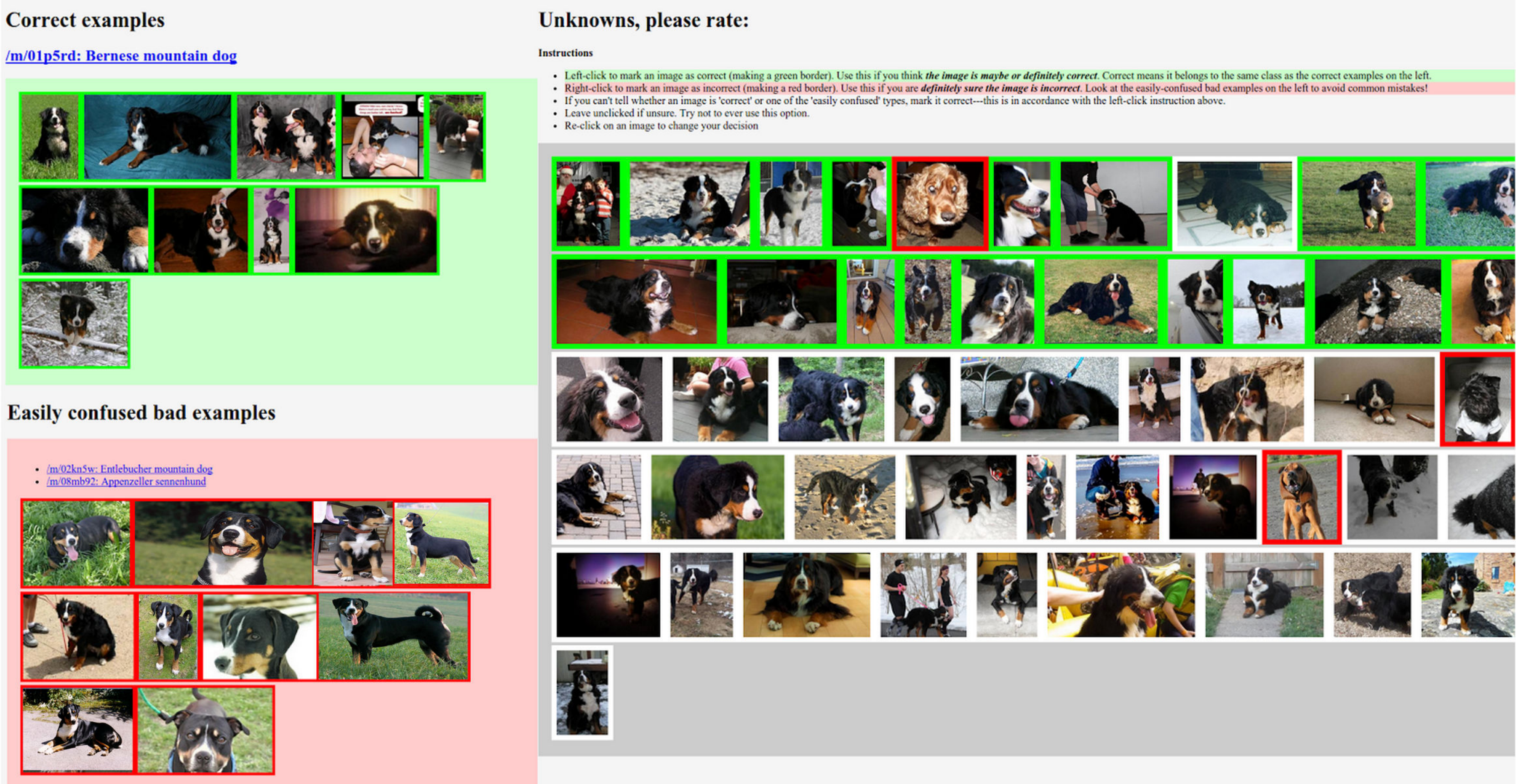}
\caption[Rater labeling tool] {
\label{fig:rater-tool2}
Our tool for binary annotation of fine-grained categories.
Instructional positive images are provided in the upper left and negatives are provided in the lower left.  This is a higher-resolution version of the figure in the main text.
}
\end{figure*}

help raters define the decision boundary between the right class and easily confused other classes.
We show the top two most confused categories, determined by the active learning's current model.
This aids in classification: in Fig.~\ref{fig:rater-tool2}, if the rater studies the positive class ``Bernese mountain dog'', they may form a mental decision rule based on fur color pattern alone.
However, when studying the negative, easily confused classes ``Entlebucher'' and ``Appenzeller'', the rater can refine the decision on more appropriate fine-grained distinctions -- in this case, hair length is a key discriminative attribute.

\subsubsection{Batching questions by class} has the benefit of allowing raters to learn about and focus on one fine-grained category at a time.
Batching questions may also allow raters to build a better mental model of the class via a human form of semi-supervised learning, although this phenomena is more difficult to isolate and measure.

\subsubsection{Golden questions for rater feedback and quality control.}
We use the original supervised seed dataset to add a number of known correct and incorrect images in the batch to be rated, which we use to give short- and long-term feedback to raters.
Short-term feedback comes in the form of a pop-up window informing the rater the moment they make an incorrect judgment, allowing them to update their mental model while working on the task.
Long-term feedback summarizes a days' worth of rating to give the rater a summary of overall performance.

\subsection{Rater Quality Improvements}
\label{app_sec:raters}

To determine the impact of our annotation framework improvements for fine-grained categories, we performed a control experiment with a more standard crowdsourcing interface, which provides only a category name, description, and image search link.
Annotation quality is determined on a set of difficult binary questions (images mistaken by a classifier on the Stanford Dogs test set).
Using our interface, annotators were both more accurate and faster, with a 16.5\% relative reduction in error (from 28.5\% to 23.8\%) and a $2.4\times$ improvement in speed (4.1 to 1.68 seconds per image).

\subsection{Annotation Statistics and Examples}
\begin{figure}[t]
\centering
\includegraphics[width=0.60\linewidth]{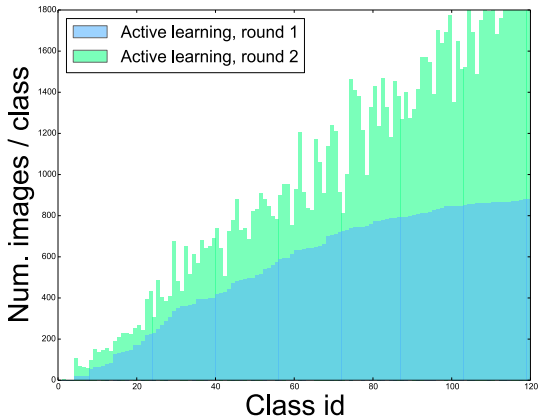}
\caption{ Counts of positive training examples obtained per category from active learning, for the Stanford Dogs dataset.}
\label{fig:al-hists}
\end{figure}

In Fig.~\ref{fig:al-hists} we show the distribution of images judged correct by human annotators after active learning selection of 1000 images per class for Stanford Dogs classes.  The categories are sorted by the number of positive training examples collected in the first iteration of active learning.  The 10 categories with the most positive training examples collected after both rounds of mining are: Pug, Golden Retriever, Boston Terrier, West Highland White Terrier, Labrador Retriever, Boxer, Maltese, German Shepherd, Pembroke Welsh Corgi, and Beagle.  The 10 categories with the fewest positive training examples are: Kerry Blue Terrier, Komondor, Irish Water Spaniel, Curly Coated Retriever, Bouvier des Flandres, Clumber Spaniel, Bedlington Terrier, Afghan Hound, Affenpinscher, and Sealyham Terrier.
These counts are influenced by the true counts of categories in the YFCC100M~\cite{thomee2015yfcc100m} dataset and our active learner's ability to find them.

\begin{figure}[t]
\centering
\includegraphics[width=0.95\linewidth]{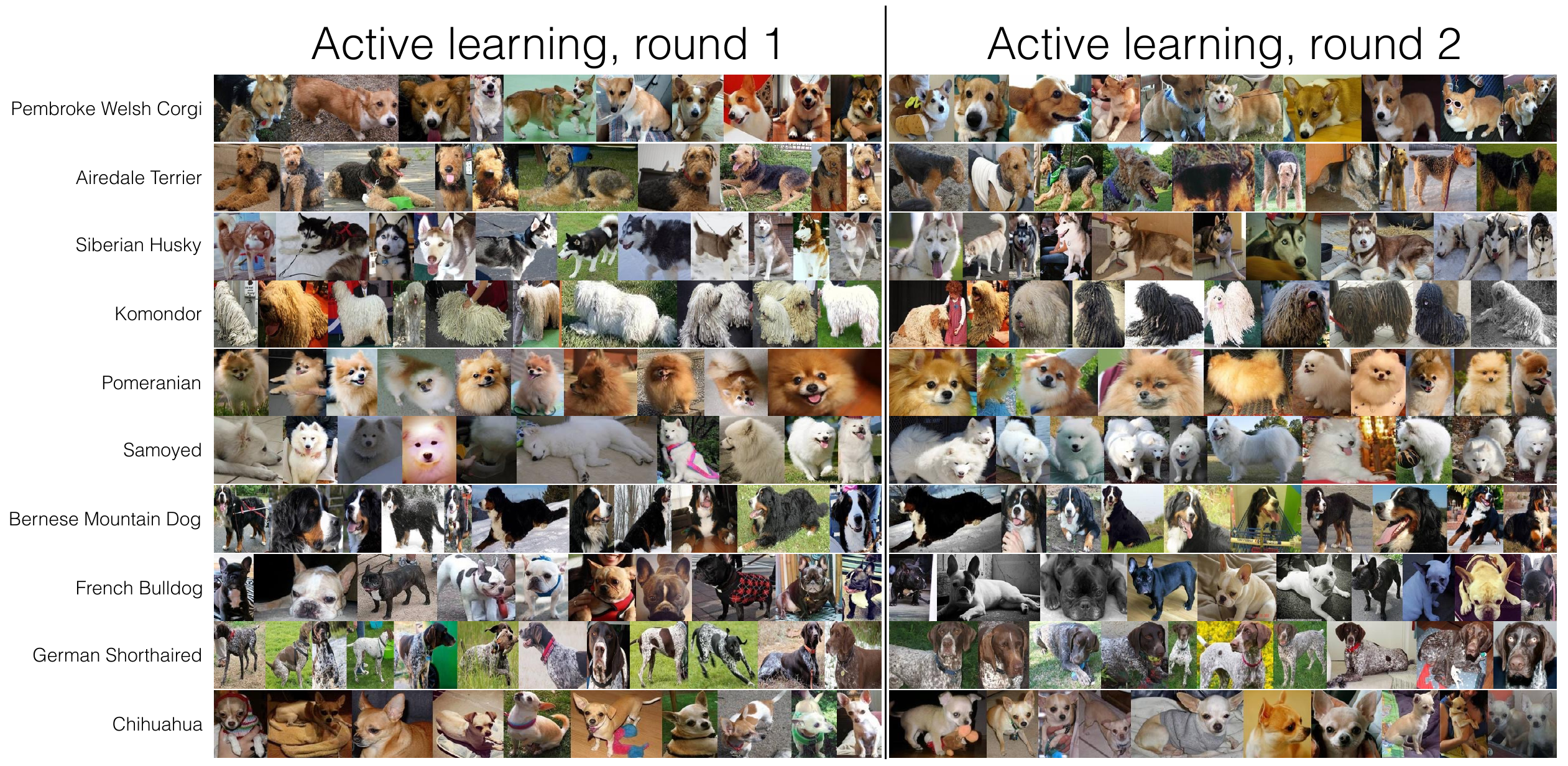}
\caption{Positive training examples obtained from active learning, from the YFCC100M dataset, for select categories from Stanford Dogs.}
\label{fig:al-qual}
\end{figure}

In Fig.~\ref{fig:al-qual}, we show positive training examples obtained from active learning for select categories, comparing examples obtained in iterations 1 and 2.

\section{Deduplication Details}
\label{app_sec:dedup}

\begin{figure}[t]
  \centering
  \begin{tabular}{ccc|ccc}
  Distance &&& Distance \tabularnewline \hline
  0 & \adjustbox{valign=m}{\includegraphics[resolution=72,scale=.2]{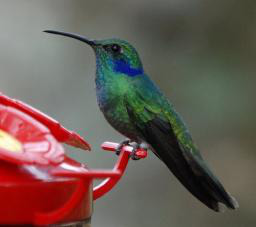}} & \adjustbox{valign=m}{\includegraphics[resolution=72,scale=.2]{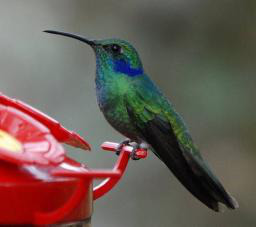}} &
  7 & \adjustbox{valign=m}{\includegraphics[resolution=72,scale=.2]{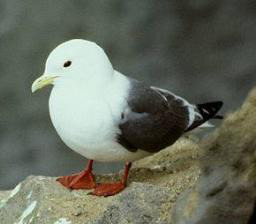}} & \adjustbox{valign=m}{\includegraphics[resolution=72,scale=.2]{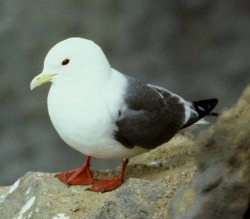}} \tabularnewline \hline
  1 & \adjustbox{valign=m}{\includegraphics[resolution=72,scale=.2]{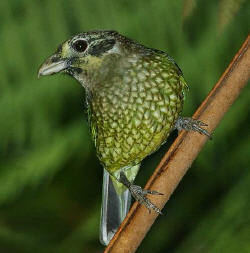}} & \adjustbox{valign=m}{\includegraphics[resolution=72,scale=.2]{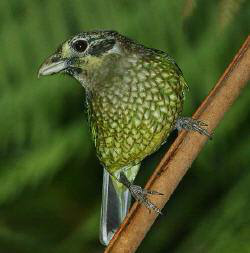}} &
  8 & \adjustbox{valign=m}{\includegraphics[resolution=72,scale=.2]{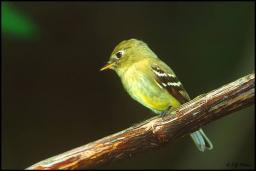}} & \adjustbox{valign=m}{\includegraphics[resolution=72,scale=.2]{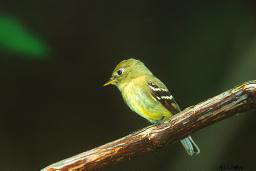}} \tabularnewline \hline
  2 & \adjustbox{valign=m}{\includegraphics[resolution=72,scale=.2]{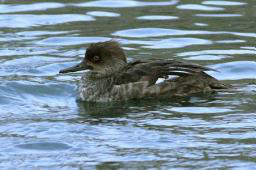}} & \adjustbox{valign=m}{\includegraphics[resolution=72,scale=.2]{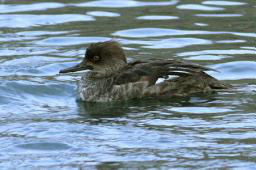}} &
  9 & \adjustbox{valign=m}{\includegraphics[resolution=72,scale=.2]{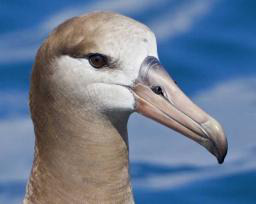}} & \adjustbox{valign=m}{\includegraphics[resolution=72,scale=.2]{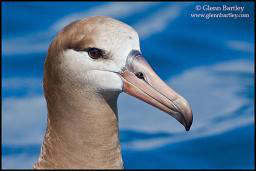}} \tabularnewline \hline
  3 & \adjustbox{valign=m}{\includegraphics[resolution=72,scale=.2]{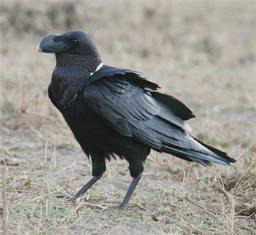}} & \adjustbox{valign=m}{\includegraphics[resolution=72,scale=.2]{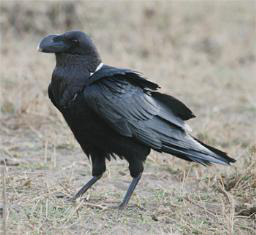}} &
 10 & \adjustbox{valign=m}{\includegraphics[resolution=72,scale=.2]{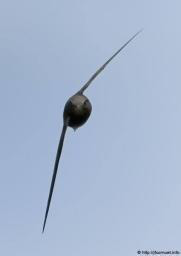}} & \adjustbox{valign=m}{\includegraphics[resolution=72,scale=.2]{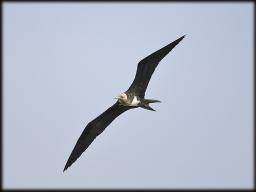}} \tabularnewline \hline
  4 & \adjustbox{valign=m}{\includegraphics[resolution=72,scale=.2]{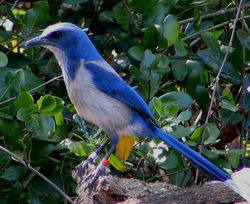}} & \adjustbox{valign=m}{\includegraphics[resolution=72,scale=.2]{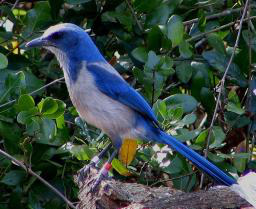}} &
 11 & \adjustbox{valign=m}{\includegraphics[resolution=72,scale=.2]{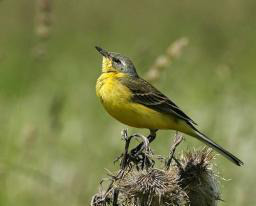}} & \adjustbox{valign=m}{\includegraphics[resolution=72,scale=.2]{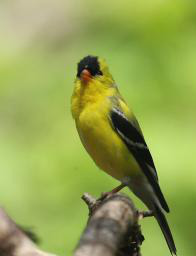}} \tabularnewline \hline
  5 & \adjustbox{valign=m}{\includegraphics[resolution=72,scale=.2]{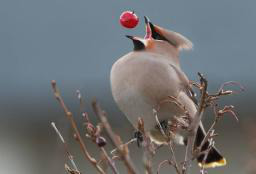}} & \adjustbox{valign=m}{\includegraphics[resolution=72,scale=.2]{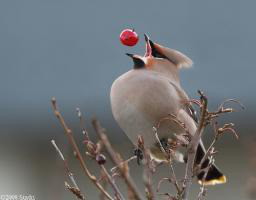}} &
 12 & \adjustbox{valign=m}{\includegraphics[resolution=72,scale=.2]{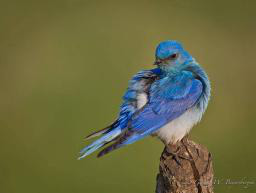}} & \adjustbox{valign=m}{\includegraphics[resolution=72,scale=.2]{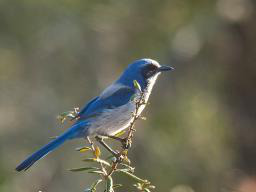}} \tabularnewline \hline
  6 & \adjustbox{valign=m}{\includegraphics[resolution=72,scale=.2]{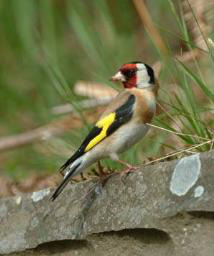}} & \adjustbox{valign=m}{\includegraphics[resolution=72,scale=.2]{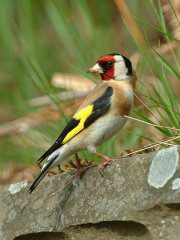}} \tabularnewline \hline
\end{tabular}
\caption{
  Example pairs of images and their distance according to our deduplication method.
  Distances 1-3 have slight pixel-level differences due to compression and the image pair at distance 4 have different scales.
  At distances 5 and 6 the images are of different crops, with distance 6 additionally exhibiting slight lighting differences.
  The images at distance 7 have slightly different scales and compression, at distance 8 there are cropping and lighting differences, and distance 9 features different crops and additional text in the corner of one photo.
  At distance 10 and higher we have image pairs which have high-level visual similarities but are distinct.
}
\label{fig:dedup}
\end{figure}

Here we provide more details on our method for removing any ground truth images from web search results, which we took great care in doing.
Our general approach follows Wang \etal~\cite{wang2014learning}, which is a state of the art method for learning a similarity metric between images.
To scale \cite{wang2014learning} to the millions of images considered in this work, we binarize the output for an efficient hashing-based exact search.
Hamming distance corresponds to dissimilarity:
identical images have distance 0, images with different resolutions, aspect ratios, or slightly different crops tend to have distances of up to roughly 4 and 8, and more substantial variations, \eg images of different views from the same photographer, or very different crops, roughly have distances up to 10, beyond which the vast majority of image pairs are actually distinct.
Qualitative examples are provided in Fig.~\ref{fig:dedup}.
We tuned our dissimilarity threshold for recall and manually verified it -- the goal is to ensure that images that have even a moderate degree of similarity to test images did not appear in our training set.
For example, of a sample of 183 image pairs at distance 16 in the large-scale bird experiments, zero were judged by a human to be too similar, and we used a still more conservative threshold of 18.
In the case of \lbird{}, 2,996 images were removed as being too similar to an image in either the CUB or Birdsnap test set.

\section{Remaining Errors: Qualitative}

\begin{figure}[t]
\centering
\includegraphics[width=.10\linewidth]{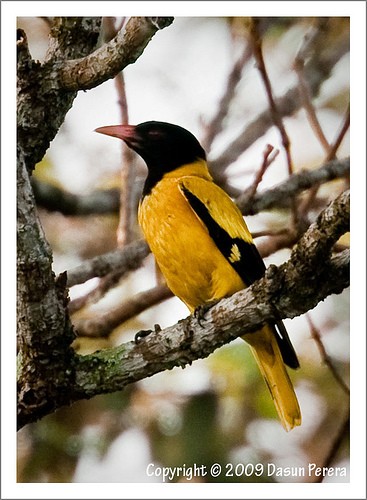}
\includegraphics[width=.10\linewidth]{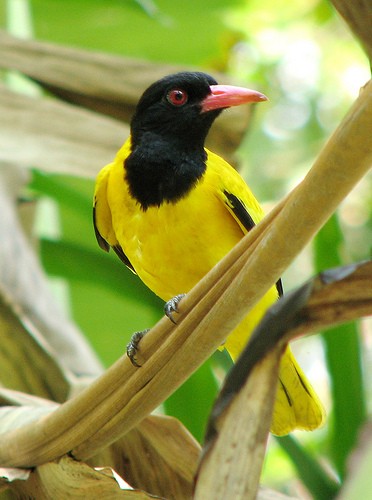}
\includegraphics[width=.10\linewidth]{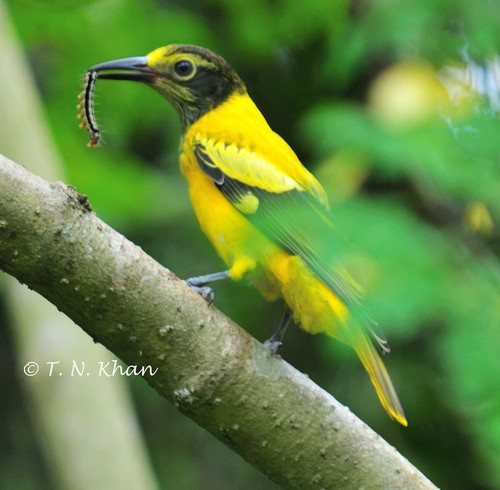}
\includegraphics[width=.10\linewidth]{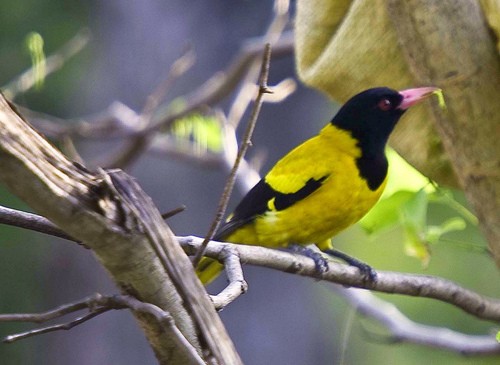}
\includegraphics[width=.10\linewidth]{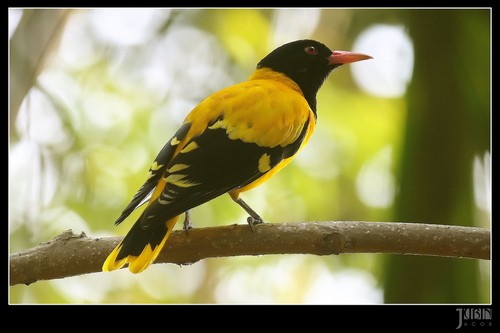}
\includegraphics[width=.10\linewidth]{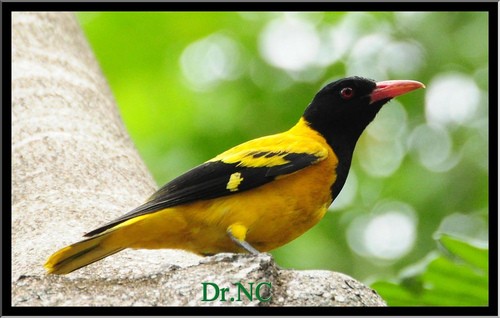}
\includegraphics[width=.10\linewidth]{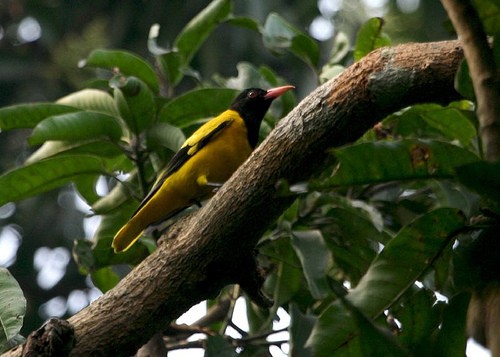}
\includegraphics[width=.10\linewidth]{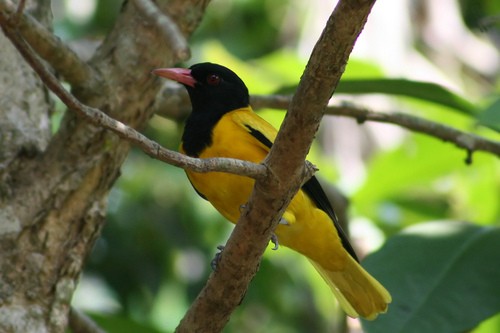}
\includegraphics[width=.10\linewidth]{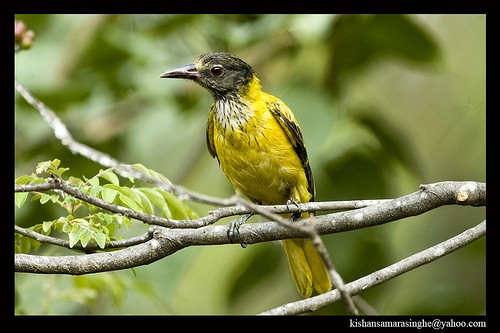}
\caption{Examples of mistakes made by a web-trained model on the CUB-200-2011~\cite{wah2011multiclass} test set, whose ground truth label is ``Hooded Oriole'', but which are actually of another species not in CUB, ``Black-Hooded Oriole.''}
\label{fig:oriole}
\end{figure}

Here we highlight one type of error that our image search model made on CUB~\cite{wah2011multiclass} -- finding errors in the test set.
We show an example in Fig.~\ref{fig:oriole}, where the true species for each image is actually a bird species not in the 200 CUB bird species.
This highlights one potential advantage of our approach: by relying on category names, web training data is tied more strongly to the semantic meaning of a category instead of simply a 1-of-$K$ label.
This also provides evidence for the ``domain shift'' hypothesis when fine-tuning on ground truth datasets, as irregularities like this can be learned, resulting in higher performance on the benchmark dataset under consideration.

\section{Network Visualization}

\begin{figure}[t]
\centering
\begin{tabular}{ll}
  \hspace{4mm} Image \hspace{5.5mm} CUB-200 \hspace{5mm} \lbird{} &
  \hspace{4mm} Image \hspace{5.5mm} CUB-200 \hspace{5mm} \lbird{}\\
  \includegraphics[width=.15\linewidth]{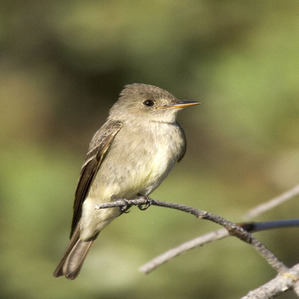}\includegraphics[width=.15\linewidth]{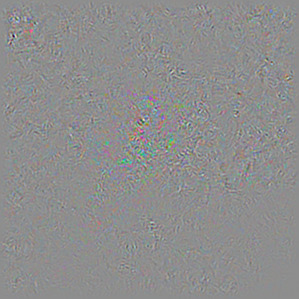}\includegraphics[width=.15\linewidth]{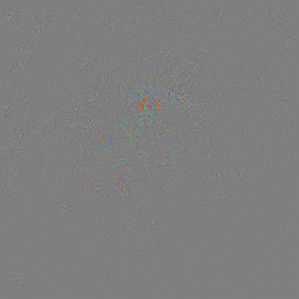} &
  \includegraphics[width=.15\linewidth]{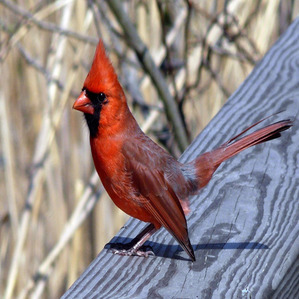}\includegraphics[width=.15\linewidth]{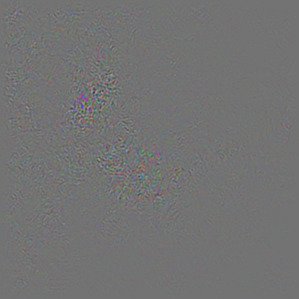}\includegraphics[width=.15\linewidth]{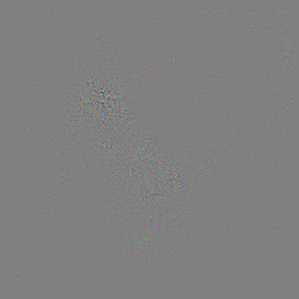} \\
  \includegraphics[width=.15\linewidth]{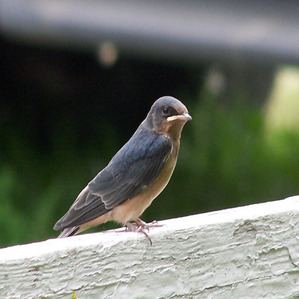}\includegraphics[width=.15\linewidth]{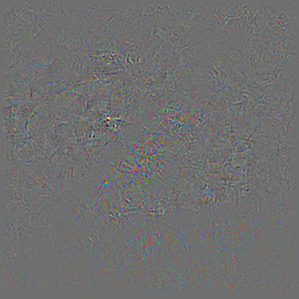}\includegraphics[width=.15\linewidth]{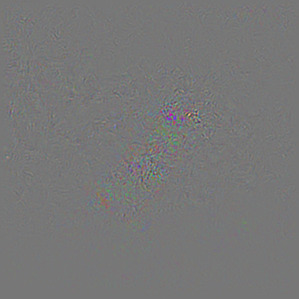} &
  \includegraphics[width=.15\linewidth]{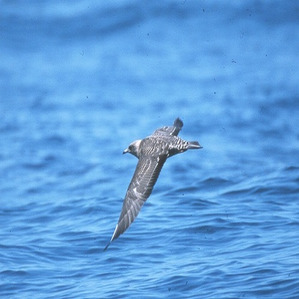}\includegraphics[width=.15\linewidth]{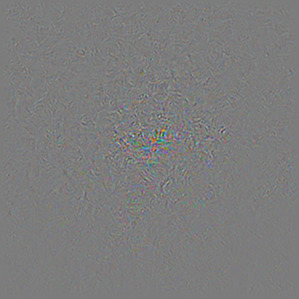}\includegraphics[width=.15\linewidth]{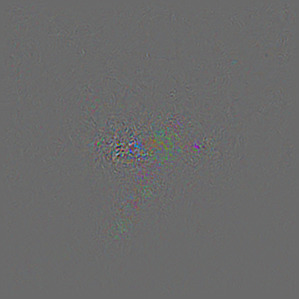} \\
  \includegraphics[width=.15\linewidth]{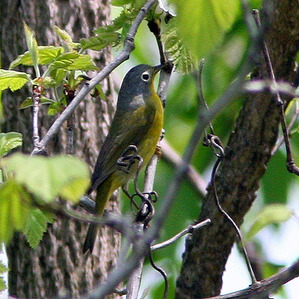}\includegraphics[width=.15\linewidth]{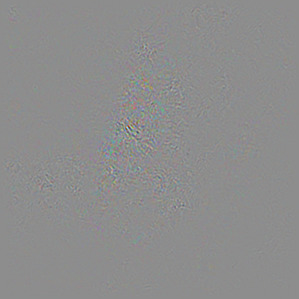}\includegraphics[width=.15\linewidth]{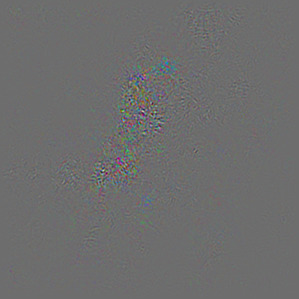} &
  \includegraphics[width=.15\linewidth]{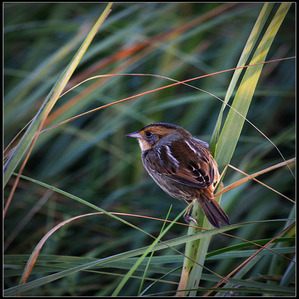}\includegraphics[width=.15\linewidth]{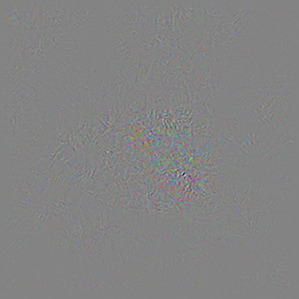}\includegraphics[width=.15\linewidth]{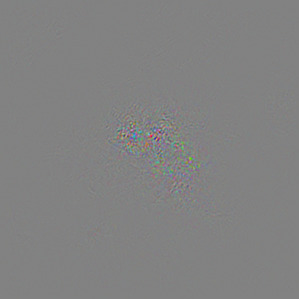} \\
  \includegraphics[width=.15\linewidth]{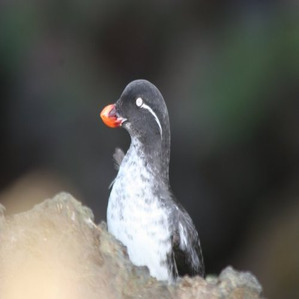}\includegraphics[width=.15\linewidth]{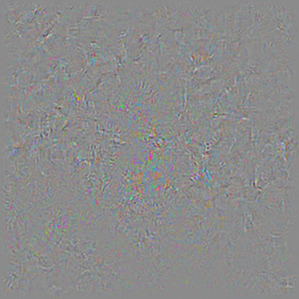}\includegraphics[width=.15\linewidth]{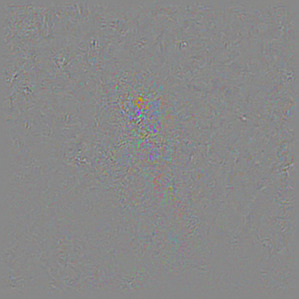} &
  \includegraphics[width=.15\linewidth]{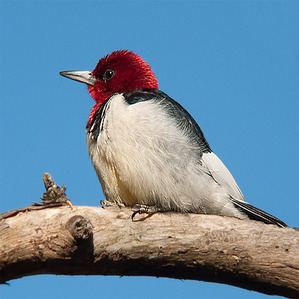}\includegraphics[width=.15\linewidth]{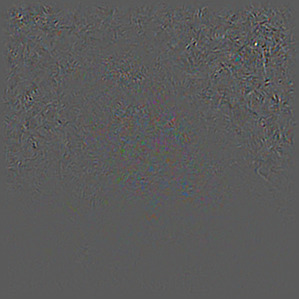}\includegraphics[width=.15\linewidth]{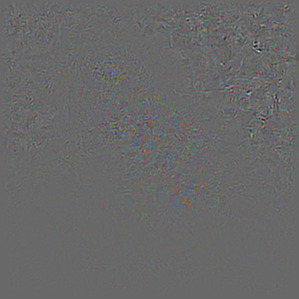} \\
  \includegraphics[width=.15\linewidth]{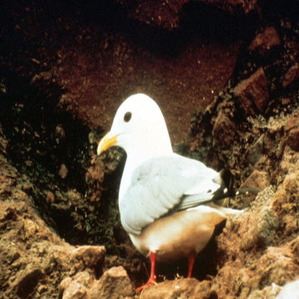}\includegraphics[width=.15\linewidth]{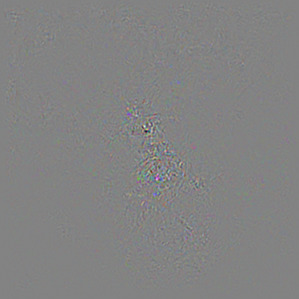}\includegraphics[width=.15\linewidth]{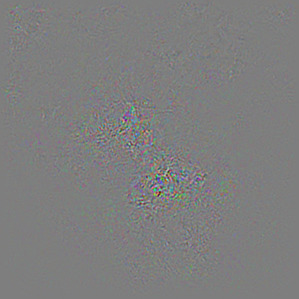} &
  \includegraphics[width=.15\linewidth]{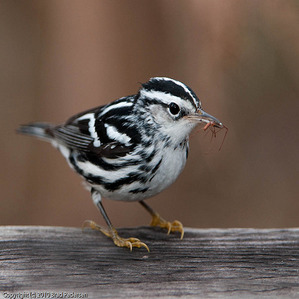}\includegraphics[width=.15\linewidth]{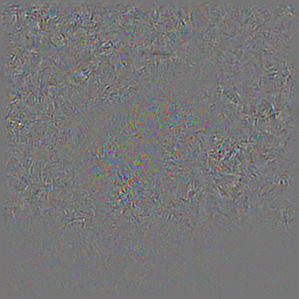}\includegraphics[width=.15\linewidth]{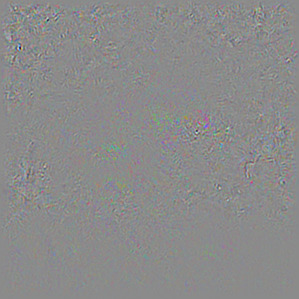} \\
\end{tabular}
\caption{
Gradients with respect to the squared norm of the last convolutional layer on ten random CUB test set images.
Each row contains, in order, an input image, gradients for a model trained on the CUB-200~\cite{wah2011multiclass} training set, and gradients for a model trained on the larger \lbird{}.
Gradients have been scaled to fit in [0,255].
Figure best viewed in high resolution on a monitor.}
\label{fig:dream}
\end{figure}

In order to examine the impact of web-trained models of fine-grained recognition from another vantage point, here we present one visualization of network internals.
Specifically, in Fig.~\ref{fig:dream} we visualize gradients with respect to the square of the norm of the last convolutional layer in the network, backpropagated into the input image, and visualized as a function of training data.
This provides some indication of the importance of each pixel with respect to the overall network activation.
Though these examples are only qualitative, we observe that the gradients for the network trained on \lbird{} are generally more focused on the bird when compared to gradients for the network trained on CUB, indicating that the network has learned a better representation of which parts of an image are discriminative.

\clearpage
\bibliographystyle{splncs03}
\bibliography{paper}

\begin{thebibliography}{10}
\providecommand{\url}[1]{\texttt{#1}}
\providecommand{\urlprefix}{URL }

\bibitem{angelova2013image}
Angelova, A., Zhu, S., Lin, Y.: Image segmentation for large-scale subcategory
  flower recognition. In: Workshop on Applications of Computer Vision (WACV).
  pp. 39--45. IEEE (2013)

\bibitem{balcan2007margin}
Balcan, M.F., Broder, A., Zhang, T.: Margin based active learning. In: Learning
  Theory, pp. 35--50. Springer (2007)

\bibitem{berg2013poof}
Berg, T., Belhumeur, P.N.: Poof: Part-based one-vs.-one features for
  fine-grained categorization, face verification, and attribute estimation. In:
  Computer Vision and Pattern Recognition (CVPR). pp. 955--962. IEEE (2013)

\bibitem{bergbirdsnapcvpr2014}
Berg, T., Liu, J., Lee, S.W., Alexander, M.L., Jacobs, D.W., Belhumeur, P.N.:
  Birdsnap: Large-scale fine-grained visual categorization of birds. In:
  Computer Vision and Pattern Recognition (CVPR) (June 2014)

\bibitem{branson2014bird}
Branson, S., Van~Horn, G., Perona, P., Belongie, S.: Improved bird species
  recognition using pose normalized deep convolutional nets. In: British
  Machine Vision Conference (BMVC) (2014)

\bibitem{branson2014ignorant}
Branson, S., Van~Horn, G., Wah, C., Perona, P., Belongie, S.: The ignorant led
  by the blind: A hybrid human--machine vision system for fine-grained
  categorization. International Journal of Computer Vision (IJCV) pp. 1--27
  (2014)

\bibitem{chai2011bicos}
Chai, Y., Lempitsky, V., Zisserman, A.: Bicos: A bi-level co-segmentation
  method for image classification. In: International Conference on Computer
  Vision (ICCV). IEEE (2011)

\bibitem{chai2013symbiotic}
Chai, Y., Lempitsky, V., Zisserman, A.: Symbiotic segmentation and part
  localization for fine-grained categorization. In: International Conference on
  Computer Vision (ICCV). pp. 321--328. IEEE (2013)

\bibitem{chai2012tricos}
Chai, Y., Rahtu, E., Lempitsky, V., Van~Gool, L., Zisserman, A.: Tricos: A
  tri-level class-discriminative co-segmentation method for image
  classification. In: European Conference on Computer Vision (ECCV), pp.
  794--807. Springer (2012)

\bibitem{chen2015webly}
Chen, X., Gupta, A.: Webly supervised learning of convolutional networks. In:
  International Conference on Computer Vision (ICCV). IEEE (2015)

\bibitem{chen2013neil}
Chen, X., Shrivastava, A., Gupta, A.: Neil: Extracting visual knowledge from
  web data. In: International Conference on Computer Vision (ICCV). pp.
  1409--1416. IEEE (2013)

\bibitem{collins2008towards}
Collins, B., Deng, J., Li, K., Fei-Fei, L.: Towards scalable dataset
  construction: An active learning approach. In: European Conference on
  Computer Vision (ECCV), pp. 86--98. Springer (2008)

\bibitem{imagenet}
Deng, J., Dong, W., Socher, R., Li, L.J., Li, K., Fei-Fei, L.: {ImageNet: A
  Large-Scale Hierarchical Image Database}. In: Computer Vision and Pattern
  Recognition (CVPR) (2009)

\bibitem{deng2013fine}
Deng, J., Krause, J., Fei-Fei, L.: Fine-grained crowdsourcing for fine-grained
  recognition. In: Computer Vision and Pattern Recognition (CVPR). pp. 580--587
  (2013)

\bibitem{divvala2014learning}
Divvala, S.K., Farhadi, A., Guestrin, C.: Learning everything about anything:
  Webly-supervised visual concept learning. In: Computer Vision and Pattern
  Recognition (CVPR). pp. 3270--3277. IEEE (2014)

\bibitem{duan2012discovering}
Duan, K., Parikh, D., Crandall, D., Grauman, K.: Discovering localized
  attributes for fine-grained recognition. In: Computer Vision and Pattern
  Recognition (CVPR). pp. 3474--3481. IEEE

\bibitem{erkan2010semi}
Erkan, A.N.: Semi-supervised learning via generalized maximum entropy. Ph.D.
  thesis, New York University (2010)

\bibitem{farrell2011birdlets}
Farrell, R., Oza, O., Zhang, N., Morariu, V.I., Darrell, T., Davis, L.S.:
  Birdlets: Subordinate categorization using volumetric primitives and
  pose-normalized appearance. In: International Conference on Computer Vision
  (ICCV). pp. 161--168. IEEE (2011)

\bibitem{fergus2010learning}
Fergus, R., Fei-Fei, L., Perona, P., Zisserman, A.: Learning object categories
  from internet image searches. Proceedings of the IEEE  98(8),  1453--1466
  (2010)

\bibitem{gavves2013fine}
Gavves, E., Fernando, B., Snoek, C.G., Smeulders, A.W., Tuytelaars, T.:
  Fine-grained categorization by alignments. In: International Conference on
  Computer Vision (ICCV). pp. 1713--1720. IEEE

\bibitem{gavves2014local}
Gavves, E., Fernando, B., Snoek, C.G., Smeulders, A.W., Tuytelaars, T.: Local
  alignments for fine-grained categorization. International Journal of Computer
  Vision (IJCV) pp. 1--22 (2014)

\bibitem{goering2014nonparametric}
Goering, C., Rodner, E., Freytag, A., Denzler, J.: Nonparametric part transfer
  for fine-grained recognition. In: Computer Vision and Pattern Recognition
  (CVPR). pp. 2489--2496. IEEE (2014)

\bibitem{he2015deep}
He, K., Zhang, X., Ren, S., Sun, J.: Deep residual learning for image
  recognition. In: Computer Vision and Pattern Recognition (CVPR). IEEE (2016)

\bibitem{Hinchliff18092015}
Hinchliff, C.E., Smith, S.A., Allman, J.F., Burleigh, J.G., Chaudhary, R.,
  Coghill, L.M., Crandall, K.A., Deng, J., Drew, B.T., Gazis, R., Gude, K.,
  Hibbett, D.S., Katz, L.A., Laughinghouse, H.D., McTavish, E.J., Midford,
  P.E., Owen, C.L., Ree, R.H., Rees, J.A., Soltis, D.E., Williams, T.,
  Cranston, K.A.: Synthesis of phylogeny and taxonomy into a comprehensive tree
  of life. Proceedings of the National Academy of Sciences  (2015),
  \url{http://www.pnas.org/content/early/2015/09/16/1423041112.abstract}

\bibitem{ioffe2015batch}
Ioffe, S., Szegedy, C.: Batch normalization: Accelerating deep network training
  by reducing internal covariate shift. In: International Conference on Machine
  Learning (ICML) (2015)

\bibitem{jaderberg2015spatial}
Jaderberg, M., Simonyan, K., Zisserman, A., Kavukcuoglu, K.: Spatial
  transformer networks. In: Neural Information Processing Systems (NIPS) (2015)

\bibitem{khosla2011novel}
Khosla, A., Jayadevaprakash, N., Yao, B., Fei-Fei, L.: Novel dataset for
  fine-grained image categorization. In: First Workshop on Fine-Grained Visual
  Categorization, Conference on Computer Vision and Pattern Recognition (CVPR).
  Colorado Springs, CO (June 2011)

\bibitem{krause2014icpr}
Krause, J., Gebru, T., Deng, J., Li, L.J., Fei-Fei, L.: Learning features and
  parts for fine-grained recognition. In: International Conference on Pattern
  Recognition (ICPR). Stockholm, Sweden (August 2014)

\bibitem{krause2015fine}
Krause, J., Jin, H., Yang, J., Fei-Fei, L.: Fine-grained recognition without
  part annotations. In: Conference on Computer Vision and Pattern Recognition
  (CVPR). IEEE

\bibitem{krause20133d}
Krause, J., Stark, M., Deng, J., Fei-Fei, L.: 3d object representations for
  fine-grained categorization. In: 4th International IEEE Workshop on 3D
  Representation and Recognition (3dRR-13). IEEE (2013)

\bibitem{kumar2012leafsnap}
Kumar, N., Belhumeur, P.N., Biswas, A., Jacobs, D.W., Kress, W.J., Lopez, I.C.,
  Soares, J.V.: Leafsnap: A computer vision system for automatic plant species
  identification. In: European Conference on Computer Vision (ECCV), pp.
  502--516. Springer (2012)

\bibitem{lecun1998gradient}
LeCun, Y., Bottou, L., Bengio, Y., Haffner, P.: Gradient-based learning applied
  to document recognition. Proceedings of the IEEE  86(11),  2278--2324 (1998)

\bibitem{lewis1994heterogeneous}
Lewis, D.D., Catlett, J.: Heterogeneous uncertainty sampling for supervised
  learning. In: International Conference on Machine Learning (ICML). pp.
  148--156 (1994)

\bibitem{li2010optimol}
Li, L.J., Fei-Fei, L.: Optimol: automatic online picture collection via
  incremental model learning. International Journal of Computer Vision (IJCV)
  88(2),  147--168 (2010)

\bibitem{mscoco}
Lin, T., Maire, M., Belongie, S., Bourdev, L.D., Girshick, R.B., Hays, J.,
  Perona, P., Ramanan, D., Doll{\'{a}}r, P., Zitnick, C.L.: Microsoft {COCO:}
  common objects in context. CoRR  abs/1405.0312 (2014),
  \url{http://arxiv.org/abs/1405.0312}

\bibitem{lin2015bilinear}
Lin, T.Y., RoyChowdhury, A., Maji, S.: Bilinear cnn models for fine-grained
  visual recognition. In: International Conference on Computer Vision (ICCV).
  IEEE

\bibitem{liu2012dog}
Liu, J., Kanazawa, A., Jacobs, D., Belhumeur, P.: Dog breed classification
  using part localization. In: European Conference on Computer Vision (ECCV),
  pp. 172--185. Springer (2012)

\bibitem{maji13finegrained}
Maji, S., Kannala, J., Rahtu, E., Blaschko, M., Vedaldi, A.: Fine-grained
  visual classification of aircraft. Tech. rep. (2013)

\bibitem{mnih2012learning}
Mnih, V., Hinton, G.E.: Learning to label aerial images from noisy data. In:
  International Conference on Machine Learning (ICML). pp. 567--574 (2012)

\bibitem{mozafari2014scaling}
Mozafari, B., Sarkar, P., Franklin, M., Jordan, M., Madden, S.: Scaling up
  crowd-sourcing to very large datasets: a case for active learning.
  Proceedings of the VLDB Endowment  8(2),  125--136 (2014)

\bibitem{nilsback2006visual}
Nilsback, M.E., Zisserman, A.: A visual vocabulary for flower classification.
  In: Computer Vision and Pattern Recognition (CVPR). vol.~2, pp. 1447--1454.
  IEEE (2006)

\bibitem{pu2014looks}
Pu, J., Jiang, Y.G., Wang, J., Xue, X.: Which looks like which: Exploring
  inter-class relationships in fine-grained visual categorization. In: European
  Conference on Computer Vision (ECCV), pp. 425--440. Springer (2014)

\bibitem{reed2014training}
Reed, S., Lee, H., Anguelov, D., Szegedy, C., Erhan, D., Rabinovich, A.:
  Training deep neural networks on noisy labels with bootstrapping. arXiv
  preprint arXiv:1412.6596  (2014)

\bibitem{russakovsky2015ilsvrc}
Russakovsky, O., Deng, J., Su, H., Krause, J., Satheesh, S., Ma, S., Huang, Z.,
  Karpathy, A., Khosla, A., Bernstein, M., Berg, A.C., Fei-Fei, L.: {ImageNet
  Large Scale Visual Recognition Challenge}. International Journal of Computer
  Vision (IJCV) pp. 1--42 (April 2015)

\bibitem{schroff2011harvesting}
Schroff, F., Criminisi, A., Zisserman, A.: Harvesting image databases from the
  web. Pattern Analysis and Machine Intelligence (PAMI)  33(4),  754--766
  (2011)

\bibitem{sermanet2014attention}
Sermanet, P., Frome, A., Real, E.: Attention for fine-grained categorization.
  arXiv preprint arXiv:1412.7054  (2014)

\bibitem{al-survey}
Settles, B.: Active learning literature survey. University of Wisconsin,
  Madison  52(55-66), ~11 (2010)

\bibitem{settles2008multiple}
Settles, B., Craven, M., Ray, S.: Multiple-instance active learning. In:
  Advances in Neural Information Processing Systems (NIPS). pp. 1289--1296
  (2008)

\bibitem{shih2015part}
Shih, K.J., Mallya, A., Singh, S., Hoiem, D.: Part localization using
  multi-proposal consensus for fine-grained categorization. In: British Machine
  Vision Conference (BMVC) (2015)

\bibitem{simon2015neural}
Simon, M., Rodner, E.: Neural activation constellations: Unsupervised part
  model discovery with convolutional networks. In: ICCV (2015)

\bibitem{simon14pdd}
Simon, M., Rodner, E., Denzler, J.: Part detector discovery in deep
  convolutional neural networks. In: Asian Conference on Computer Vision
  (ACCV). vol.~2, pp. 162--177 (2014)

\bibitem{sukhbaatar2014learning}
Sukhbaatar, S., Fergus, R.: Learning from noisy labels with deep neural
  networks. arXiv preprint arXiv:1406.2080  (2014)

\bibitem{szegedy2016inception}
Szegedy, C., Ioffe, S., Vanhoucke, V.: Inception-v4, inception-resnet and the
  impact of residual connections on learning. arXiv preprint arXiv:1602.07261
  (2016)

\bibitem{szegedy2014going}
Szegedy, C., Liu, W., Jia, Y., Sermanet, P., Reed, S., Anguelov, D., Erhan, D.,
  Vanhoucke, V., Rabinovich, A.: Going deeper with convolutions. In: Computer
  Vision and Pattern Recognition (CVPR) (2015)

\bibitem{szegedy2015rethinking}
Szegedy, C., Vanhoucke, V., Ioffe, S., Shlens, J., Wojna, Z.: Rethinking the
  inception architecture for computer vision. In: Computer Vision and Pattern
  Recognition (CVPR). IEEE (2016)

\bibitem{thomee2015yfcc100m}
Thomee, B., Shamma, D.A., Friedland, G., Elizalde, B., Ni, K., Poland, D.,
  Borth, D., Li, L.J.: The new data and new challenges in multimedia research.
  arXiv preprint arXiv:1503.01817  (2015)

\bibitem{torralba2011unbiased}
Torralba, A., Efros, A., et~al.: Unbiased look at dataset bias. In: Computer
  Vision and Pattern Recognition (CVPR). pp. 1521--1528. IEEE (2011)

\bibitem{horn2015}
Van~Horn, G., Branson, S., Farrell, R., Haber, S., Barry, J., Ipeirotis, P.,
  Perona, P., Belongie, S.: Building a bird recognition app and large scale
  dataset with citizen scientists: The fine print in fine-grained dataset
  collection. In: Computer Vision and Pattern Recognition (CVPR). IEEE (2015)

\bibitem{mahendran14understanding}
Vedaldi, A., Mahendran, S., Tsogkas, S., Maji, S., Girshick, B., Kannala, J.,
  Rahtu, E., Kokkinos, I., Blaschko, M.B., Weiss, D., Taskar, B., Simonyan, K.,
  Saphra, N., Mohamed, S.: Understanding objects in detail with fine-grained
  attributes. In: Computer Vision and Pattern Recognition (CVPR) (2014)

\bibitem{wahcub2002011}
Wah, C., Branson, S., Welinder, P., Perona, P., Belongie, S.: {The Caltech-UCSD
  Birds-200-2011 Dataset}. Tech. Rep. CNS-TR-2011-001, California Institute of
  Technology (2011)

\bibitem{wah2013attribute}
Wah, C., Belongie, S.: Attribute-based detection of unfamiliar classes with
  humans in the loop. In: Computer Vision and Pattern Recognition (CVPR). pp.
  779--786. IEEE (2013)

\bibitem{wah2011multiclass}
Wah, C., Branson, S., Perona, P., Belongie, S.: Multiclass recognition and part
  localization with humans in the loop. In: International Conference on
  Computer Vision (ICCV). pp. 2524--2531. IEEE (2011)

\bibitem{wah2014similarity}
Wah, C., Horn, G., Branson, S., Maji, S., Perona, P., Belongie, S.: Similarity
  comparisons for interactive fine-grained categorization. In: Computer Vision
  and Pattern Recognition (CVPR) (2014)

\bibitem{wang2014learning}
Wang, J., Song, Y., Leung, T., Rosenberg, C., Wang, J., Philbin, J., Chen, B.,
  Wu, Y.: Learning fine-grained image similarity with deep ranking. In:
  Proceedings of the IEEE Conference on Computer Vision and Pattern
  Recognition. pp. 1386--1393 (2014)

\bibitem{welinderetal2010}
Welinder, P., Branson, S., Mita, T., Wah, C., Schroff, F., Belongie, S.,
  Perona, P.: {Caltech-UCSD Birds 200}. Tech. Rep. CNS-TR-2010-001, California
  Institute of Technology (2010)

\bibitem{xiao2015application}
Xiao, T., Xu, Y., Yang, K., Zhang, J., Peng, Y., Zhang, Z.: The application of
  two-level attention models in deep convolutional neural network for
  fine-grained image classification. In: Computer Vision and Pattern
  Recognition (CVPR). IEEE

\bibitem{xiao2015learning}
Xiao, T., Xia, T., Yang, Y., Huang, C., Wang, X.: Learning from massive noisy
  labeled data for image classification. In: Computer Vision and Pattern
  Recognition (CVPR). IEEE

\bibitem{xie2014hyper}
Xie, S., Yang, T., Wang, X., Lin, Y.: Hyper-class augmented and regularized
  deep learning for fine-grained image classification. In: Computer Vision and
  Pattern Recognition (CVPR). IEEE

\bibitem{xu2015augmenting}
Xu, Z., Huang, S., Zhang, Y., Tao, D.: Augmenting strong supervision using web
  data for fine-grained categorization. In: International Conference on
  Computer Vision (ICCV) (2015)

\bibitem{yang2015large}
Yang, L., Luo, P., Loy, C.C., Tang, X.: A large-scale car dataset for
  fine-grained categorization and verification. In: Computer Vision and Pattern
  Recognition (CVPR). IEEE

\bibitem{yang2012unsupervised}
Yang, S., Bo, L., Wang, J., Shapiro, L.G.: Unsupervised template learning for
  fine-grained object recognition. In: Advances in Neural Information
  Processing Systems (NIPS). pp. 3122--3130 (2012)

\bibitem{yao2012codebook}
Yao, B., Bradski, G., Fei-Fei, L.: A codebook-free and annotation-free approach
  for fine-grained image categorization. In: Computer Vision and Pattern
  Recognition (CVPR). pp. 3466--3473. IEEE (2012)

\bibitem{yao2011combining}
Yao, B., Khosla, A., Fei-Fei, L.: Combining randomization and discrimination
  for fine-grained image categorization. In: Computer Vision and Pattern
  Recognition (CVPR). pp. 1577--1584. IEEE (2011)

\bibitem{yu2015construction}
Yu, F., Zhang, Y., Song, S., Seff, A., Xiao, J.: Construction of a large-scale
  image dataset using deep learning with humans in the loop. arXiv preprint
  arXiv:1506.03365  (2015)

\bibitem{zhang2014part}
Zhang, N., Donahue, J., Girshick, R., Darrell, T.: Part-based r-cnns for
  fine-grained category detection. In: European Conference on Computer Vision
  (ECCV), pp. 834--849. Springer (2014)

\bibitem{zhang2012pose}
Zhang, N., Farrell, R., Darrell, T.: Pose pooling kernels for sub-category
  recognition. In: Computer Vision and Pattern Recognition (CVPR). pp.
  3665--3672. IEEE (2012)

\bibitem{zhang2013deformable}
Zhang, N., Farrell, R., Iandola, F., Darrell, T.: Deformable part descriptors
  for fine-grained recognition and attribute prediction. In: International
  Conference on Computer Vision (ICCV). pp. 729--736. IEEE (2013)

\bibitem{zhang2015weakly}
Zhang, Y., Wei, X.s., Wu, J., Cai, J., Lu, J., Nguyen, V.A., Do, M.N.: Weakly
  supervised fine-grained image categorization. arXiv preprint arXiv:1504.04943
   (2015)

\end{thebibliography}
\end{document}